\documentclass[runningheads]{llncs}
\usepackage{graphicx}
\usepackage{comment}
\usepackage{amsmath,amssymb} 
\usepackage{color}

\usepackage[utf8]{inputenc} 
\usepackage[T1]{fontenc}    
\usepackage{url}            
\usepackage{booktabs}       
\usepackage{amsfonts}       
\usepackage{nicefrac}       
\usepackage{microtype}      
\usepackage[dvipsnames]{xcolor}
\usepackage[backref=page,hyperindex,breaklinks]{hyperref}
\usepackage{graphicx}
\usepackage{amsmath}
\usepackage{bm}
\usepackage{dsfont}
\usepackage{paralist}

\usepackage{soul}
\usepackage[normalem]{ulem}
\usepackage{caption} 
\usepackage{setspace}

\usepackage{subfigure}
\usepackage{wrapfig}
\usepackage{epsfig}
\usepackage{multirow}
\usepackage{wrapfig}
\usepackage{mathtools,cuted}

\usepackage{xcolor,amsmath}
\usepackage{pifont}
\usepackage{algorithm,algpseudocode}
\usepackage{bbold}

\input{taesup.sty}

\newcommand{\cmark}{\ding{51}}%
\newcommand{\xmark}{\ding{55}}%

\newcommand{\eg}{{\it e.g.}}

\newcommand{\tsmcolor}{black}
\usepackage{xr}

\definecolor{commentcolor}{RGB}{000,000,204}   
\newcommand{\PyComment}[1]{\ttfamily\textcolor{commentcolor}{\# #1}}  
\newcommand{\PyCode}[1]{\ttfamily\textcolor{black}{#1}} 

\newlength\myindent
\usepackage{authblk}
\usepackage{footnote}
\newcommand*\samethanks[1][\value{footnote}]{\footnotemark[#1]}

\makeatletter
\newcommand{\settitle}{\@maketitle}
\makeatother

\makeatletter
\newcommand{\maketitlepage}{%
  \begin{titlepage}
    \let\thanks\@gobble
    \let\footnote\@gobble
    \if@twocolumn
      \ifnum \col@number=\@ne
        \@maketitle
      \else
        \twocolumn[\@maketitle]%
      \fi
    \else
      \@maketitle
    \fi
    \thispagestyle{empty}
  \end{titlepage}
}
\makeatother
\author{Jaeseok Byun\inst{1}\thanks{Equal contribution. This work was performed when Jaeseok Byun did an internship at Microsoft Research Asia.} \and Taebaek Hwang\inst{2}\samethanks \and Jianlong Fu\inst{3} \and Taesup Moon\inst{1}\thanks{Corresponding author (\texttt{tsmoon@snu.ac.kr})}}
\institute{Department of ECE/ASRI/IPAI, Seoul National University
\and Department of ECE, Sungkyunkwan University \and Microsoft Research Asia  \\
\email{wotjr3868@snu.ac.kr, gxq9106@gmail.com, jianf@microsoft.com, tsmoon@snu.ac.kr }
}
\begin{document}
\pagestyle{headings}
\mainmatter
\title{GRIT-VLP: Grouped Mini-batch Sampling \\for Efficient Vision and Language Pre-training} 
\titlerunning{GRIT-VLP: Grouped Mini-batch Sampling for Efficient VLP}
\authorrunning{J.Byun et al.}
\maketitle
\begin{abstract}
Most of the currently existing vision and language pre-training (VLP) methods have mainly focused on how to extract and align vision and text features. 
In contrast to the mainstream VLP methods, we highlight that two routinely applied steps during pre-training have crucial impact on the performance of the pre-trained model: \textit{in-batch} hard negative sampling for image-text matching (ITM) and assigning the large masking probability for the masked language modeling (MLM).
After empirically showing the unexpected effectiveness of above two steps, we systematically devise our GRIT-VLP, which adaptively samples mini-batches for more effective mining of hard negative samples for ITM while maintaining the computational cost for pre-training. Our method consists of three components:
1) GRouped mIni-baTch sampling (GRIT) strategy that collects similar examples in a mini-batch, 2) ITC consistency loss for improving the mining ability, and 3) enlarged masking probability for MLM.
Consequently, we show our GRIT-VLP achieves a new state-of-the-art performance on various downstream tasks with much less computational cost. Furthermore, we demonstrate that our model is essentially in par with ALBEF, the previous state-of-the-art, only with one-third of training epochs on the same training data. Code is available at \href{https://github.com/jaeseokbyun/GRIT-VLP}{https://github.com/jaeseokbyun/GRIT-VLP}.
\keywords{Efficient vision and language pre-training, hard negative sampling, batch-sampling strategy, shuffling}
\end{abstract}
\section{Introduction}
 Recently, the pre-training and fine-tuning approach of the Transformer \cite{(attention)vaswani2017attention} based models have made exciting progress in natural-language-processing (NLP) \cite{(BERT)devlin2018bert} and vision tasks \cite{(VIT)dosovitskiy2020image}. Particularly, despite the huge computational cost, vision and language pre-training (VLP)  \cite{(ALBEF)li2021align,(LXMERT)tan2019lxmert,(SOHO)huang2021seeing,(CLIP)radford2021learning,(UNITER)chen2020uniter,(ViLT)kim2021vilt,(OSCAR)li2020oscar,(VL-BERT)su2019vl,(Visualbert)li2019visualbert,(VILLA)gan2020large}, which aims to learn cross-modal representations from large-scale image-text pairs, enabled to achieve the state-of-the-art results in various vision and language downstream tasks, \eg, image-text retrieval (IRTR), natural language for visual reasoning (NLVR) \cite{(NLVR)suhr2018corpus}, and visual question answering (VQA) \cite{(VQA)antol2015vqa}, etc.
For the joint understanding of image and text, a multi-modal encoder used in VLP is typically trained with the self-supervised learning objectives, such as image-text matching (ITM) and masked language modeling (MLM). 

Majority of the existing VLP methods have focused on how to make the vision features to align with those of the text.
The first popular approach \cite{(UNITER)chen2020uniter,(LXMERT)tan2019lxmert,(VilBERT)lu2019vilbert,(OSCAR)li2020oscar} is to utilize the salient region-based features extracted from a pre-trained object detector. However, these region feature based VLP methods suffer from severe computational inefficiency and heavy dependency on the pre-trained object detectors. 
In order to overcome such drawbacks, recent approaches have replaced the object detectors with CNN backbones \cite{(PixelBERT)huang2020pixel,(SOHO)huang2021seeing} or linear embedding inspired by the recently developed Vision Transformer (ViT) \cite{(VIT)dosovitskiy2020image}, which enables efficient end-to-end training of the  vision-language representation learning.

Recently, ALBEF \cite{(ALBEF)li2021align} was proposed as another attempt to lift the dependency on the object detectors. They designed a novel VLP architecture to integrate the uni-modal encoder for each modality (\textit{i.e.,} an object-detector-free vision encoder and a text encoder) by employing a multi-modal Transformer encoder that fuses features from them. Additionally, ALBEF employed the image-text contrastive (ITC) loss for uni-modal encoders to \textit{pre-align} the features before fusing, the \textit{in-batch} hard negative sampling strategy for the ITM, and a momentum distillation 
to further improve the performance. As a result, it achieved the state-of-the-art performance for the multiple vision and language downstream tasks. 

While the main emphasis of \cite{(ALBEF)li2021align} was on the pre-aligning stage via ITC, we double-check that proposition and carry out careful ablation analyses on ALBEF and identify that the two routinely applied sampling steps in fact have crucial impacts on the final downstream performance. 
\begin{figure*}[t]
    \centering
    \subfigure{
    \includegraphics[width=\linewidth]{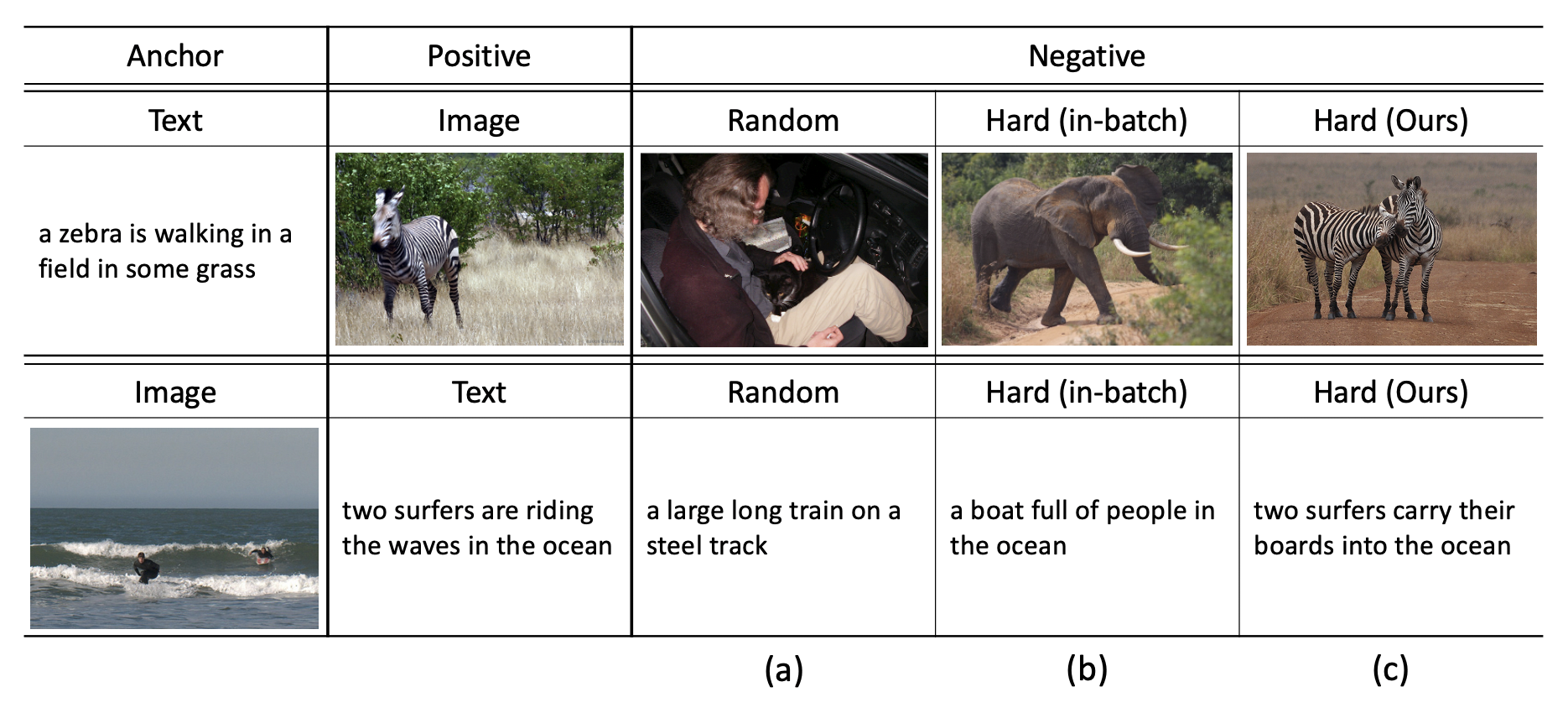}}
    \caption{A comparison of negative samples for ITM task selected by (a) Random, (b) \textit{In-batch} sampling by ALBEF  \cite{(ALBEF)li2021align}, and (c) Our GRIT strategy. }
    \label{fig:samples_for_itm}
\begin{spacing}{0.1}
\end{spacing}
\end{figure*}
Firstly, the \textit{hard negative sampling} for the ITM task, of which effect was described as marginal compared to the pre-aligning in \cite[Section 6]{(ALBEF)li2021align},
in fact turns out to be an essential component, even more than the ITC itself, for efficient VLP. Namely, when training for the ITM task, rather than using the randomly selected negatives as shown in Fig. \ref{fig:samples_for_itm}(a) (for the text and image anchor, respectively), selecting \textit{hard} negative samples as in Fig. \ref{fig:samples_for_itm}(b), which is sampled from the given \textit{mini-batch} by using the contrastive image-text similarity scores already computed for the ITC, becomes much more useful for promoting a more fine-grained representation learning. Secondly, the \textit{mask sampling probability} for the MLM task, which typically is naively set to $15\%$, also is shown to have a significant impact on the VLP performance. Namely, when the probability is enlarged up to 50\%, the multi-modal encoder is enforced to use more visual context features for predicting the masked token, hence, a more consolidated multi-modal representations could be learned. A recent concurrent work \cite{(Masking40)wettig2022should} also suggests the enlarging of the masking probability for MLM; however, their focus was on the NLP domain, thus, they have not investigated the impact of enlarging the probability on the multi-modal encoder.

Motivated by above analyses, in this paper, we make the following three modifications on ALBEF to significantly improve the downstream performance and computational efficiency of VLP. First, we devise GRIT (GRouped mIni-baTch sampling) strategy that enables to select much more informative hard negative samples (as shown in Fig. \ref{fig:samples_for_itm}(c)) than those in \cite{(ALBEF)li2021align} (Fig. \ref{fig:samples_for_itm}(b)), \textit{without} introducing any significant memory and computational overhead. Note such improvement is far from being straightforward since a naive extension of previous approaches would require either additional GPU memory (when simply enlarging the batch size) or forward pass computation (when utilizing additional queues as in \cite{(hard_metric_cross_batch)wang2020cross,(MOCO)he2020momentum,(memorybank)wu2018unsupervised}). We elaborate on this point more in details in a later section. Second, we devise a consistency loss between the image-text similarity scores used for ITC such that the contrastive learning and pre-aligning become more effective and, as a result, enables our GRIT to sample more exquisite negative samples. Third, we use enlarged mask sampling probability (50\%) for MLM such that the visual features can be further integrated with the text features when solving the downstream tasks. 

Our final method that combines above modifications is dubbed as GRIT-VLP, and we show that it can significantly improve the efficiency of VLP compared to ALBEF. Namely, trained on the exact same training data, GRIT-VLP significantly outperforms ALBEF on all of the downstream tasks we tested with 33\% fewer number of epochs, 21\% less training time per epoch. Furthermore, our thorough analyses show that GRIT-VLP is model agnostic and can be easily applied to existing VLP with different model architectures and objectives, which demonstrates the potential of our method being an essential tool for VLP. 

\vspace{-.1in}
\section{Preliminaries and Related Work}
\noindent\textbf{[Vision-language pre-training]} 
Existing VLP methods, which can be categorized into three frameworks, have mainly focused on the development of objectives and architectures to learn multi-modal representations.
The first approach is to adopt dual uni-modal encoders which are composed of separate image and text encoder.
CLIP \cite{(CLIP)radford2021learning} and ALIGN \cite{(ALIGN)jia2021scaling} pre-trained with contrastive learning have been shown to be effective for IRTR, without object detectors.
However, they suffer from the performance degradation in other downstream tasks (\eg, VQA, NLVR).
The second approach \cite{(LXMERT)tan2019lxmert,(VilBERT)lu2019vilbert,(VL-BERT)su2019vl,(OSCAR)li2020oscar,(Unicoder-VL)li2020unicoder} mainly utilizes a single multi-modal encoder where concatenated text and image representations are used as input.
In contrast to the former approach, these works consistently show promising results on various downstream tasks.
However, these methods heavily depend on the pre-trained object detectors which are computationally inefficient. 
Thus, recent works \cite{(PixelBERT)huang2020pixel,(SOHO)huang2021seeing,(ViLT)kim2021vilt,(Probing)xue2021probing} have struggled to replace object detectors with more efficient ones.
The last category \cite{(ALBEF)li2021align,(TCL)yang2022vision} offsets the shortcomings of the previous approaches by combining them, and achieves state-of-the-art performance. 
ALBEF \cite{(ALBEF)li2021align} combines them by adding pre-alignmenet before fusing. 
Our method is built upon this ALBEF \cite{(ALBEF)li2021align}, but, deviating from the mainstream of VLP, our attention is on the \textit{sampling strategy} for efficient pre-training. 

\noindent\textbf{[Hard negative mining]} 
Most prior works on negative mining \cite{(hard_meric_sample)wu2017sampling,(hard_metric_smart)harwood2017smart,(Facenet)schroff2015facenet,(useful_hard)xuan2020hard,(Easy_positive)xuan2020improved} point out that hard negatives can help a training model to converge faster. 
Recent approaches  \cite{(Contrastive_hard)robinson2020contrastive,(Debiased_hard)chuang2020debiased,(False_neg_conditional)wu2020conditional,(False_neg_incremental)chen2021incremental,(CrossCLR)zolfaghari2021crossclr} mainly focus on the unsupervised contrastive learning setting where true dissimilarity of pairs are not available. 
However, these methods can not be applied to the VLP methods (second, third categories in the previous paragraph) due to the inherent architecture and input of multi-modal encoder.

\subsection{ALign BEfore Fuse (ALBEF) \cite{(ALBEF)li2021align}}
Since ALBEF is the base model on which we build our method, we review it in details here. 
It consists of an image encoder $f_v$, a text encoder $f_t$, and a multi-modal encoder $h$, all of which are based on the Transformer architecture. 
Each input image $V$ and sentence $T$ is encoded into respective embedding sequences: $f_v(V)=\{v^{cls},v^{1},v^{2},...,v^{S_V}\}$ and $f_t(T)=\{t^{cls},t^{1},t^{2},...,t^{S_T}\}$, in which $v^{cls}$ and $t^{cls}$ denote the embedding of the [CLS] token for each modality, and $S_V$ and $S_T$ denote the sequence length of image and text, respectively. 
Then, vision and text representations are fused by a cross-attention module in the multi-modal encoder which requires both vision and text features as input (\textit{i.e.}, $h(f_v(V),f_t(T))= \{w^{cls},w^{1},w^{2},...,w^{S_T}\}$). 
The three pre-training objectives of ALBEF are briefly introduced below\footnote{Note the Momentum Distillation (MD), which utilizes the soft outputs from an additional momentum model is omitted, since we do NOT use the momentum model.}. 

\noindent{\textbf{(a) Image-text contrastive learning (ITC) }} focuses on the pre-alignment of uni-modal representations before fusing them with a multi-modal encoder. 
Like conventional contrastive learning, it promotes positive image-text pairs to have similar representations and negative ones to be dissimilar. 
Inspired by MoCo \cite{(MOCO)he2020momentum}, ALBEF utilizes two \textit{queues} for storing recent [CLS] embeddings from the unimodal encoders, \textit{i.e.,} $v^{cls}$ and $t^{cls}$, and use them as extra negatives for the contrastive learning. 
More specifically, a similarity between $V$ and $T$ is defined as $s(V,T) = g_v(v^{cls})^{T}g_t(t^{cls})$ in which $g_v(\cdot)$ and $g_t(\cdot)$ are linear projections for mapping [CLS] embeddings to the normalized lower dimensional features.
Then, for each $V$ and $T$, the normalized image-to-text and text-to-image similarities for $j=1,\ldots,N$ are defined as: 
\begin{equation}
    p^{v2t}_{j}(V) = \frac{\exp(s(V,T_j)/\tau)}{\sum_{j=1}^{N}\exp(s(V,T_j)/\tau)}, \,\,\,
    p^{t2v}_{j}(T) = \frac{\exp(s(V_j,T)/\tau)}{\sum_{j=1}^{N}\exp(s(V_j,T)/\tau)}, \label{eq:itc_similarity}
\end{equation}
in which $\tau$ is a learnable temperature, and $N$ is the size of the queue. 
The ITC loss is then defined as:
\begin{equation}
   \mathcal{L}_\text{ITC} = \frac{1}{2}\mathbb{E}_{(V,T) \sim D} [\texttt{CE}( \bm{y}^{v2t}(V),\bm{p}^{v2t}(V) ) + \texttt{CE}( \bm{y}^{t2v}(T),\bm{p}^{t2v}(T) ) ], \label{eq:ITC_loss}
\end{equation}
in which $\bm{y}^{v2t}(V)$ and $\bm{y}^{t2v}(T)$ denotes the ground-truth one-hot vector for the true pair sample for $V$ and $T$, respectively. Now, in the pre-training, we do \textit{not} use the queues for ITC but use the \textit{in-batch} version, \textit{i.e.,} $N$ in (\ref{eq:itc_similarity}) is the size of the mini-batch, to implement a lightweight version in terms of memory/computation. 

\noindent{\textbf{(b) Image-text matching (ITM)}} is a binary classification task that predicts whether a pair of image and text, $(V,T)$, is matched or not. The prediction probability of the classifier, $\bm{p}^\text{ITM}(V,T)$, is obtained by using the joint embedding feature of [CLS] token ($w^{cls}$) from the multi-modal encoder.
Then, the ITM loss is defined as 
\begin{equation}
   \mathcal{L}\textsubscript{ITM} = \mathbb{E}_{(V,T) \sim D} [\texttt{CE}( \bm{y}^{\text{ITM}},\bm{p}^\text{ITM}(V,T))]. \label{eq:ITM_loss}
\end{equation}
in which $\bm{y}^{ITM}$ is the ground truth one-hot vector, and 
$\texttt{CE}(\cdot, \cdot)$ stands for the cross-entropy between the two probability vectors. The effectiveness of ITM is determined by the quality of the negative pair, and, as outlined in the Introduction, ALBEF proposes the \textit{in-batch} hard negative sampling (ITM\textsubscript{hard}) by utilizing $\bm p^{v2t}(V)$ and $\bm p^{t2v}(T)$ defined in (\ref{eq:itc_similarity}) for sampling text and image that has high similarity for given $V$ and $T$, respectively, as a negative sample pair. 

\noindent{\textbf{(c) Masked language modeling (MLM)}} is a task to predict the randomly masked tokens in a text based on both contextual text and visual information. 
ALBEF uses the masking probability of $15\%$ following \cite{(BERT)devlin2018bert}, and by denoting the randomly masked text as $\tilde{T}$ and the prediction probability for the masked tokens as $\bm{p}^\text{mask}(V,\tilde{T})$, the loss function of MLM becomes
\begin{equation}
   \mathcal{L}_\text{MLM} = \mathbb{E}_{(V,\tilde{T}) \sim D} [\texttt{CE}( \bm{\tilde{y}},\bm{p}^\text{mask}(V,\tilde{T}))],           \label{eq:MLM_loss}
\end{equation}
in which $\tilde{\bm y}$ is a ground truth one-hot vector for the masked token. 
\begin{figure*}[t]
    \centering
    \subfigure{
    \includegraphics[width=\linewidth]{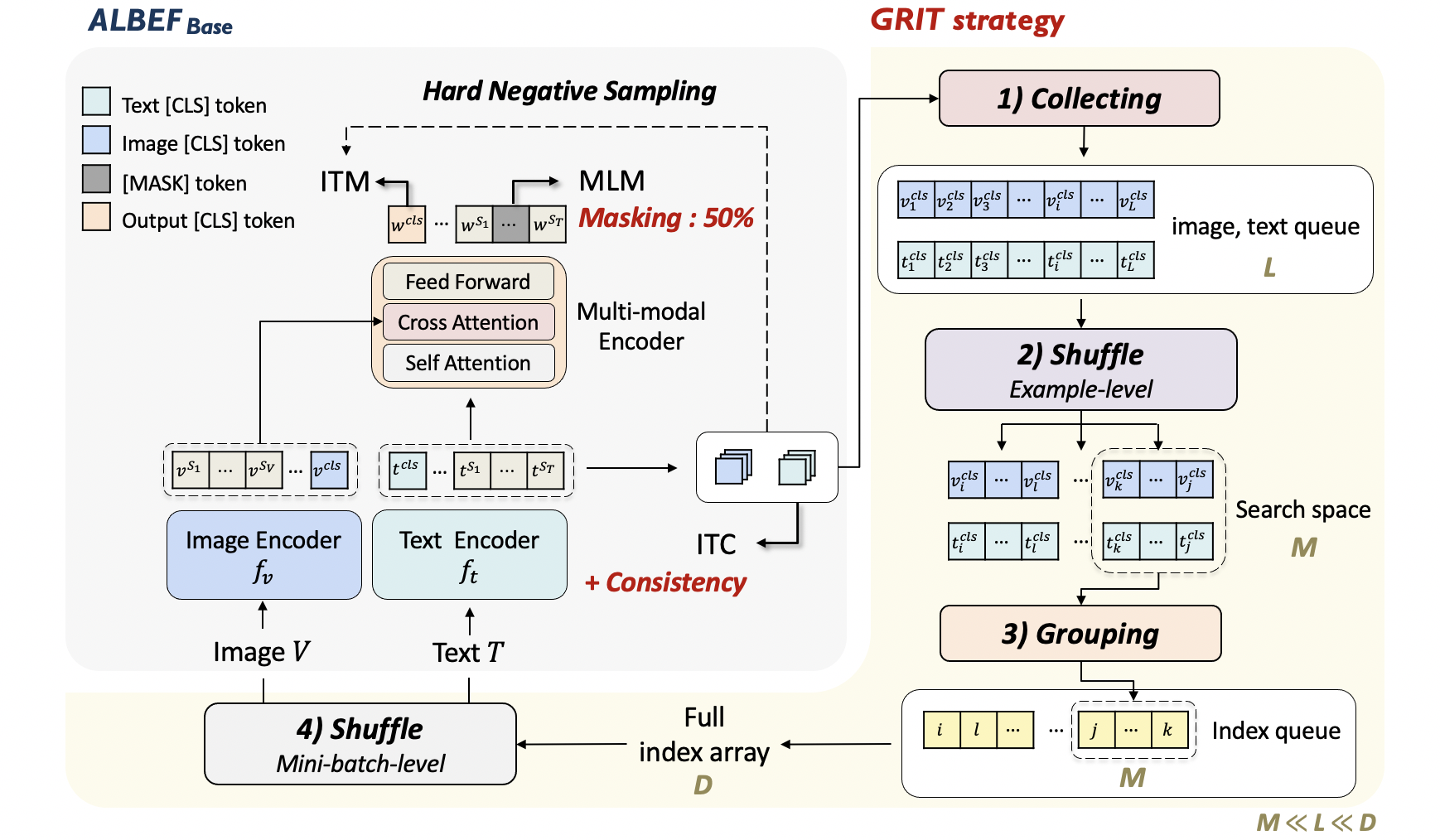}}
    \caption{The ALBEF\textsubscript{Base} architecture and the overall process of GRIT-VLP.} 
    \label{fig:overall_process}
\begin{spacing}{-0.2}
\end{spacing}
\end{figure*}

\vspace{-.1in}
\section{Ablation Analyses on ALBEF} \label{sec:background}
As mentioned in the Introduction, we carry out careful analyses on ALBEF to verify the true effect of the training objectives described in the previous section. To that end, we set the base model as ``ALBEF\textsubscript{Base}'', which mostly follows the model architecture and pre-training objectives 
of ALBEF, but does \textit{not} use the additional momentum encoder and momentum distillation\footnote{We defer describing the detailed model architecture to Section \ref{subsec:data_exp}.}. Highlighting the training objectives, we can also denote ``ALBEF\textsubscript{Base}'' by MLM+ITM\textsubscript{hard}+ITC, and we ablate each of those components and evaluate the performance of the model variants on two downstream tasks (IRTR, NLVR). All models are pre-trained with 4M dataset, and evaluated with MS-COCO \cite{(MS-COCO)lin2014microsoft} and NLVR2 dataset \cite{(NLVR)suhr2018corpus}. 
Details on the tasks, datasets and additional results are described in Section \ref{sec:experiments} and the Supplementary Material (S.M.).\\

\noindent{\textbf{[Hard negative sampling on ITM]}} 
Table \ref{tab:itm_analysis} compares the downstream task performance of models that have the fixed MLM objective (with masking probability 15\%) but varying ITM and ITC objectives of ALBEF\textsubscript{Base}. In the table,  ``MLM+ITM\textsubscript{hard}'' stands for the case in which only ITM\textsubscript{hard} is carried out without the ITC objective --- this case is missing in the analysis of the original ALBEF paper \cite[Table 1]{(ALBEF)li2021align}, but we believe it is necessary for showing the effect of ITM\textsubscript{hard} alone without the pre-algining effect of ITC. The subtlety here is that, since ITM\textsubscript{hard} utilizes the image-text similarity scores from ITC (\ref{eq:itc_similarity}) for selecting the \textit{in-batch} hard negative samples, we use the scores obtained from the uni-modal encoders of ALBEF (without the multi-modal encoder) that are \textit{pre-trained} only with the ITC loss. Moreover, ``ITM\textsubscript{rand}'' in Table 1 stands for the ITM loss with randomly selected negative samples. 
\begin{table}[!t]
\centering
\caption{Ablation study on ITM\textsubscript{hard} and ITC for ALBEF\textsubscript{Base}. 
} 
\label{tab:itm_analysis}
\resizebox{\linewidth}{!}{%
\begin{tabular}{lllllll|lclclclclclcllclcl} 
\hline
\multirow{2}{*}{\begin{tabular}[c]{@{}l@{}}\\\end{tabular}} & \multirow{2}{*}{Epochs} & \multirow{2}{*}{} & \multirow{2}{*}{} & \multirow{2}{*}{Training tasks} & \multirow{2}{*}{} & \multirow{2}{*}{} & \multirow{2}{*}{} &               &  & TR            &  & \multicolumn{3}{c}{(COCO)}  &  & IR            &  & \multicolumn{1}{l}{} &  & \multicolumn{1}{c}{} & \multicolumn{3}{c}{NLVR}  &   \\
                                                            &                         &                   &                   &                                 &                   &                   &                   & R@1           &  & R@5           &  & R@10          &  & R@1           &  & R@5           &  & R@10                 &  &                      & (val) &  & (test)         &   \\ 
\hline

                                                            & \multirow{4}{*}{10}     & \multirow{4}{*}{} & \multirow{4}{*}{}

                                                            &  MLM + ITM\textsubscript{rand}                       &                   &                   &                   & 61.6          &  & 86.1          &  & 92.5          &  & 47.8          &  & 75.4          &  & 84.8                 &  &                      & 77.02 &  & 78.44          &   \\
                                                            &                         &                   &                   & MLM + ITM\textsubscript{rand} + ITC                        &                   &                   &                   & 66.8          &  & 88.8          &   & 94.5          &  & 51.1          &  & 78.4          &  & 86.8                 &  &                      & 76.59 &  & 78.69          &   \\
                                                            &                         &                   &                   & MLM + ITM\textsubscript{hard}                    &                   &                   &                   & 68.6          &  & 89.4          &  & 94.9          &  & 52.1          &  & 79.0          &  & 87.1                 &  &                      & 79.18 &  & 79.32          &   \\
                                                            
                                                            &                         &                   & 
                                                            & ALBEF\textsubscript{Base}               &                   &                   &                   & 72.3          &  & 91.3          &  & 96.0          &  & 55.1          &  & 81.0          &  & 88.5                 &  &                      & 79.21 &  & 79.78          &   \\
                                                            
\hline

                                                            & \multirow{4}{*}{20}     & \multirow{4}{*}{} & \multirow{4}{*}{} & MLM + ITM\textsubscript{rand}                       &                   &                   &                   & 66.5          &  & 88.3          &  & 94.0          &  & 51.3          &  & 78.3          &  & 86.5                 &  &                      & 78.02 &  & 79.43          &   \\
                                                            &                         &                   &                   & MLM + ITM\textsubscript{rand} + ITC                 &                   &                   &                   & 69.6          &  & 90.9          &  & 95.3          &  & 53.8          &  & 80.0          &  & 87.8                 &  &                      & 77.61 &  & 79.43          &   \\
                                                            &                         &                   &                   & MLM + ITM\textsubscript{hard}                   &                   &                   &                   & 72.0          &  & 91.5          &  & \textbf{96.6}          &  & 57.5          &  & 81.2          &  & 88.4                 &  &                      & \textbf{80.44} &  & \textbf{80.83} &   \\
                                                            &                         &                   &                   & ALBEF\textsubscript{Base}            &                   &                   &                   & \textbf{73.8} &  & \textbf{92.3} &  & 96.5 &  & \textbf{57.7} &  & \textbf{82.5} &  & \textbf{89.6}        &  &                      & 79.22 &  & 80.37          &   \\
\hline

\end{tabular}
}
\begin{spacing}{-0.2}
\end{spacing}
\end{table}

The original ALBEF essentially focuses on the effect of ITC by mainly comparing ``MLM+ITM\textsubscript{rand}'' and ``MLM+ITM\textsubscript{rand}+ITC'' and argues that ``ITM\textsubscript{hard}'' only gives a marginal improvement when it replaces ITM\textsubscript{rand}. However, we observe a different story in Table \ref{tab:itm_analysis}. Namely, even without the pre-alignment of the representations via ITC, ``MLM+ITM\textsubscript{hard}'' gives a significant performance boost over ``MLM+ITM\textsubscript{rand}'', which is substantially larger than the improvement we get by ``MLM+ITM\textsubscript{rand}+ITC''. Moreover, even with a shorter 10 epochs, ``MLM+ITM\textsubscript{hard}'' performs competitively or superior to ``MLM+ITM\textsubscript{rand}+ITC'' trained for a longer 20 epochs. While the best performance is still obtained by using ITM\textsubscript{hard} and ITC together, \textit{i.e.}, ALBEF\textsubscript{Base}, this result strongly motivates that further improving ITM\textsubscript{hard} could be central in attaining efficient VLP.  \\

\noindent{\textbf{[Mask sampling probability for MLM]}} 
Table \ref{tab:mlm_analysis} now focuses on MLM by varying the masking probability with fixed ITM and ITC. Namely, the original ALBEF\textsubscript{Base} trains with the masking probability of 15\%, and we also test the model with the probability 50\%, dubbed as ALBEF\textsubscript{Base$_{50}$}. In the table, we observe that this simple change brings surprising performance gain; ALBEF\textsubscript{Base$_{50}$} always outperforms ALBEF\textsubscript{Base} for the same epoch and becomes comparable to ALBEF\textsubscript{Base} even when trained with significantly smaller number of epochs. This result clearly motivates using enlarged masking probability for MLM for VLP. 

\subsection{Motivation}

The result in Table \ref{tab:itm_analysis} suggests that improving the hard negative sampling strategy for ITM could bring further performance gain for VLP. An obvious way for such improvement is to enlarge the search space from which the negative samples are selected, hence, the sample that contains the nuanced difference with respect to the positive sample as in Figure \ref{fig:samples_for_itm}(c) can be obtained. However, we note that such enlargement in a memory- and computation-efficient way is far from being straightforward, described as below. 
\begin{table}[!t]
\centering
\caption{Ablation study on the masking probability for MLM for ALBEF\textsubscript{Base}.
}
\label{tab:mlm_analysis}
\resizebox{\linewidth}{!}{%
\begin{tabular}{lllllll|llclclclclclcll|lclcl} 
\hline

\multirow{2}{*}{\begin{tabular}[c]{@{}l@{}}\\~\\~\\\end{tabular}} & \multirow{2}{*}{Epochs} & \multirow{2}{*}{} & \multirow{2}{*}{} & \multirow{2}{*}{Training tasks} & \multirow{2}{*}{} & \multirow{2}{*}{} & \multirow{2}{*}{} & \multirow{2}{*}{} &                &                      & TR             &                      & \multicolumn{3}{c}{(COCO)}                             &  & IR             &                      & \multicolumn{1}{l}{} &                      &                       &                      & \multicolumn{3}{c}{NLVR}                                 &   \\
                                                                  &                         &                   &                   &                                 &                   &                   &                   &                   & R@1            &                      & R@5            & \multicolumn{3}{c}{R@10}                                     & R@1            &  & R@5            & \multicolumn{3}{c}{R@10}                                           &                       &                      & (val)           &                      & (test)          &   \\ 
\midrule
                                                                  & \multirow{2}{*}{10}     & \multirow{4}{*}{} & \multirow{4}{*}{} & ALBEF\textsubscript{Base}                       &                   &                   &                   &                   & 72.3           &                      & 91.3           &                      & 96.0           &                      & 55.1           &  & 81.0           &                      & 88.5                 &                      &                       &                      & 79.21           &                      & 79.78           &   \\
                                                                  &                         &                   &                   & ALBEF\textsubscript{Base$_{50}$}                     &                   &                   &                   &                   & \textbf{73.4 } & \multicolumn{1}{c}{} & \textbf{92.5 } & \multicolumn{1}{c}{} & \textbf{96.4 } & \multicolumn{1}{c}{} & \textbf{57.2 } &  & \textbf{82.3 } & \multicolumn{1}{c}{} & \textbf{89.4 }       & \multicolumn{1}{c}{} & \multicolumn{1}{c|}{} & \multicolumn{1}{c}{} & \textbf{79.42 } & \multicolumn{1}{c}{} & \textbf{79.87 } &   \\
\hline

                                                                  & \multirow{2}{*}{20}     & \multirow{4}{*}{} & \multirow{4}{*}{} & ALBEF\textsubscript{Base}                       &                   &                   &                   &                   & 73.8           &                      & 92.3           &                      & 96.5           &                      & 57.7           &  & 82.5           &                      & 89.6                 &                      &                       &                      & 79.22           &                      & 80.37           &   \\
                                                                  &                         &
                                                                  &              &ALBEF\textsubscript{Base$_{50}$}                    &                   &                   &                   &                   & \textbf{75.6}  &                      & \textbf{93.2}  &                      & \textbf{96.7}  &                      & \textbf{58.8}  &  & \textbf{83.2}  &                      & \textbf{90.1}        &                      &                       &                      & \textbf{80.41}           &                      & \textbf{80.54}  &   \\
                                                      
\hline
\end{tabular}
}
\begin{spacing}{0.1}
\end{spacing}
\end{table}

The most naive way to enlarge the search space  is to enlarge the size of the mini-batch during training. While conceptually simple, it clearly is limited by the GPU memory and high computational cost. 
 An alternative is to utilize the additional queues to store the compressed representations of the samples (\textit{i.e.,} the [CLS] tokens [$v^{cls},t^{cls}$] from the uni-modal encoders), like MoCo  \cite{(MOCO)he2020momentum} or MemoryBank \cite{(memorybank)wu2018unsupervised}, and include those representations in the search space for mining the hard negatives. While this queue-based solution is highly effective in the ordinary contrastive learning, it causes additional complication for VLP using the ITM loss. Namely, as described in Section \ref{sec:background}, the ITM loss is calculated with the [CLS] token from the multi-modal encoder ($w^{cls}$), which needs the \textit{entire} sequence embeddings ($f_v(V)$, $f_t(T)$) to compute. Therefore, to employ the queue-based solution for ITM\textsubscript{hard}, 
 one should select between the following two options. One is to store the entire embedding sequences for both modalities in the queues, which is severely memory-inefficient due to the long sequence lengths (typically, $S_T: $ $30$ and $S_V: $ $200$). The other is to only store [$v^{cls},t^{cls}$] tokens from each modality to compute (\ref{eq:itc_similarity}) and (\ref{eq:ITC_loss}) for ITC, but carry out the additional forward passes for the samples that are not in the current mini-batch to compute $w^{cls}$ and the ITM loss. Clearly, the second option would suffer from the additional computation cost required for the forward passes. 

To overcome the limitations of the above naive solutions, we propose a new method that can enlarge the search space and select more informative negative samples for ITM \textit{without} any significant overheads on the memory/computation. 
\section{Main Method: GRIT-VLP}\label{sec:methods}
\subsection{GRouped mIni-baTch sampling (GRIT)}
\begin{figure*}[t]
    \centering
    \subfigure{
    \includegraphics[width=\linewidth]{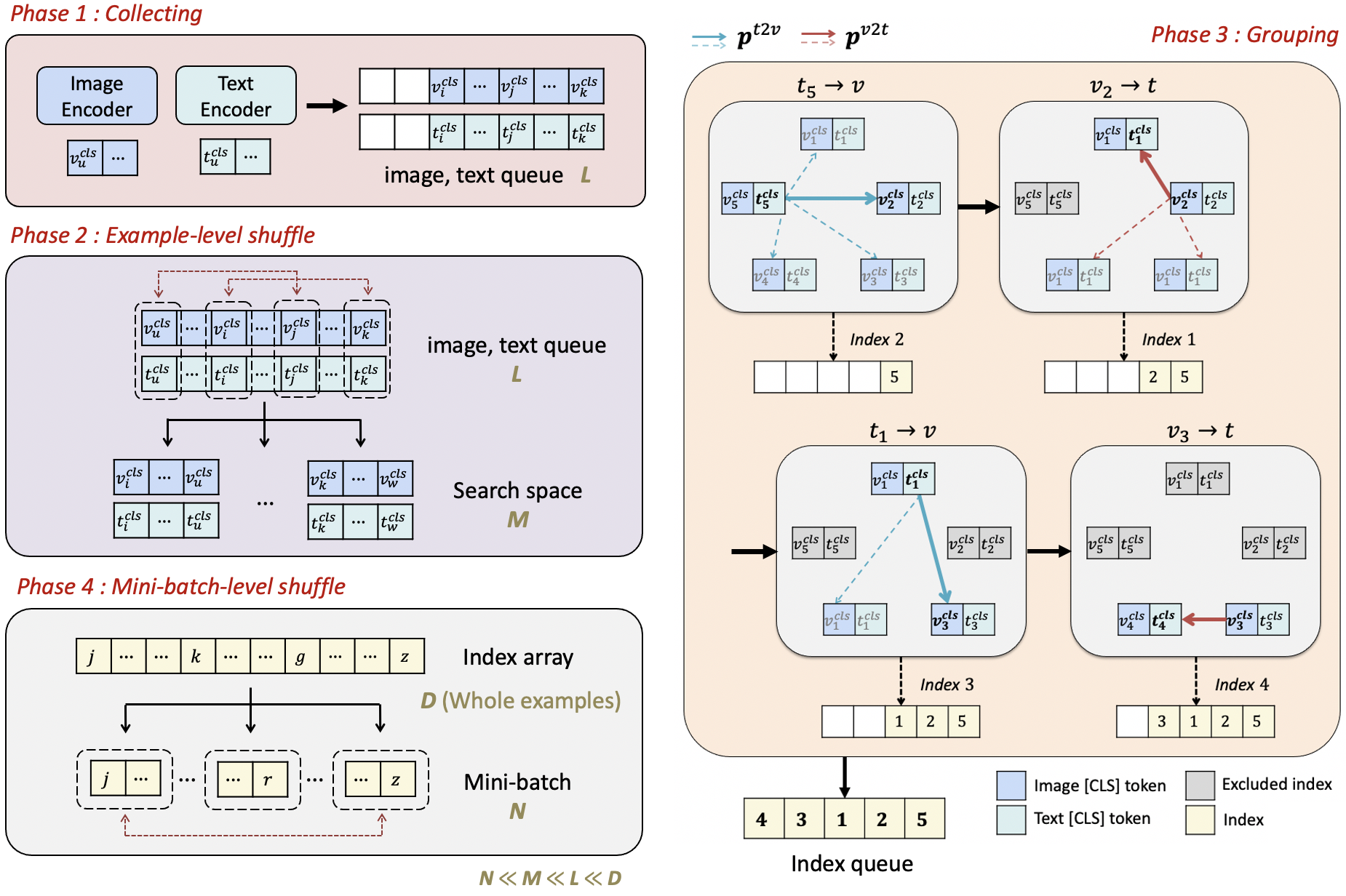}}
     \caption{Grouped mini-batch sampling (GRIT)  strategy.}
    \label{fig:GMS_shuffle}
\begin{spacing}{0.1}
\end{spacing}
\end{figure*}
In this section, we describe our main contribution, the GRouped mIni-baTch sampling (GRIT) strategy. 
The basic idea is to compose each mini-batch of size $N$ with highly similar example groups such that the informative, hard negative samples are likely to be chosen by the \textit{in-batch} sampling of ALBEF. In order to do that without significant memory/computation overhead, as described in Fig. \ref{fig:overall_process} and Algorithm 1 in S.M, GRIT utilizes two additional queues of size $L$ that store the [CLS] tokens from the uni-modal encoders, an index queue $I$ of size $M$, and a full index array $G$ of size $D$ (the whole data size). The ordering of the sizes is $N \ll M \ll L \ll D$. Then, the procedure of constructing \textit{grouped} mini-batches for the next epoch is performed concurrently with the loss calculation for pre-training at each epoch, and these grouped mini-batches are used for the ordinary mini-batch training in the following epoch. 

A subtle challenge of above simultaneous creation process for the grouped mini-batches is that it removes the randomness of the mini-batches, which is an essential ingredient for the \textit{stochastic} gradient descent based learning. 
Therefore, we add two-level shuffling phases for preserving the randomness among the grouped mini-batches.
As a result, our GRIT is composed of the following four phases: 1) collecting, 2) example-level shuffling 3) grouping, and 4) mini-batch-level shuffling. We note the first three phases are repeated whenever the queue of size $L$ is filled, and the last phase is repeated once every epoch. 

\noindent{\textbf{[\textit{Phase 1}: Collecting]}}
To construct mini-batches containing similar samples, we first store the [CLS] tokens $[v^{cls},t^{cls}]$ (from uni-modal encoders) in the two additional queues of size $L$, which is much larger than the size of mini-batches ($N$), until filled as shown in Figure \ref{fig:GMS_shuffle}. 

\noindent{\textbf{[\textit{Phase 2}: Example-level shuffle]}}
Once the queues are filled, all the samples in the queues are randomly shuffled at the example-level, to secure the randomness mentioned above. 
After shuffling, each queue is divided into $\frac{L}{M}$ sub-queues of size $M$, which is the size of the enlarged search space for the hard negative samples for ITM. Then, the samples in each sub-queue are grouped based on similarity via the grouping phase below, which is sequentially performed for each sub-queue. 

\noindent{\textbf{[\textit{Phase 3}: Grouping]}}
From the $[v^{cls},t^{cls}]$ stored in the sub-queue, we can compute the image-to-text and text-to-image similarity scores, similarly as in (\ref{eq:itc_similarity}), among the examples in the sub-queue. Accordingly, for each pair $(V,T)$, those scores can be denoted by $\bm q^{v2t}(V)\in\Delta^M$ and $\bm q^{t2v}(T)\in\Delta^M$, respectively. 

Based on the computed similarities, we aim group \textit{similar} $(V,T)$ examples in the sub-queue to each mini-batch as much as possible. To that end, 
as described in Algorithm 2 in S.M, our grouping phase is summarized as:
1) randomly sample the first pair $(V_1,T_1)$ from the sub-queue, then 2) iteratively find and store the index of the most similar example one by one until all examples inside the sub-queue are visited once, and finally, 3) the index queue $I \in \{1,\ldots,M\}^M$ is generated.
Note both the negative text for an anchor image and the negative image for an anchor text should be considered when constructing the negative samples for ITM. 
Thus, rather than using a one-way similarity score, two similarity scores are used alternatively; namely, as illustrated in Fig. \ref{fig:GMS_shuffle} with a toy example of $M=5$, at the $(i+1)$-th iteration, given a pair $(V_k,T_k)$ with index $k$, $I_{i+1}$ is chosen as
\begin{equation}
      I_{i+1} =
    \begin{cases}
       \argmax_{j \notin I} \bm{q}^{t2v}_{j}(T_k) & \text{if   } I_{i} \text{  is chosen with   }  \bm{q}^{v2t}\\
     \argmax_{j \notin I} \bm{q}^{v2t}_{j}(V_k) & \text{if   } I_{i} \text{  is chosen with   }  \bm{q}^{t2v}.\\
    \end{cases}     
\label{eq:grouping}
\end{equation}
Thus, during above \textit{grouping} process for the sub-queue, half of the pairs are selected based on (V $\rightarrow$ T) direction, and the other half based on (T $\rightarrow$ V) direction. 
Whenever the index queue $I$ is full, we convert the indices into the original data indices in $\{1,\ldots,D\}$
and append those to the full index array $G$. 

\noindent{\textbf{[\textit{Phase 4}: Mini-batch-level shuffle]}}
After each epoch, the full index array $G$, which is a permutation of   $\{1,\ldots,D\}$, is generated. Then, $G$ is divided into multiple mini-batch-sized arrays, and these arrays are shuffled. 
Note this shuffling is done at the mini-batch level, not at the example level. Finally, these shuffled mini-batches are used for both training and GRIT for the next epoch. \\

\noindent\textit{Remark 1:} We note the shuffling phases \textit{Phase 2/4} in GRIT are important to secure the randomness among the mini-batches. Namely, since GRIT generates the indices during the previous epoch, it omits the conventional data re-shuffling performed at the start of each epoch. Hence, although the order of indices is continuously changed to some extent in \textit{Phase 3}, such re-ordering happens only at the level of sub-queue of size $M$, hence the scope of shuffling is significantly limited. 
In Table \ref{tab:ablation} (Section \ref{sec:experiments}), we verify that the performance of GRIT without shuffling is significantly degraded, justifying the proposed shuffling phases.

\noindent\textit{Remark 2: } 
The naive implementation of GRIT would be to proceed \textit{Phase 1/3} and training \textit{separately}, not concurrently. To be specific, at the beginning of each epoch, the conventional re-shuffling of the whole data is done, followed by additional forward passes on the uni-modal encoders, and \textit{Phase 1/3} are performed to generate grouped mini-batch indices. Then, the training begins with the generated indices. Since this naive version requires additional forward passes, it clearly has high computational cost and requires longer training time. 

\vspace{-.1in}
\subsection{ITC consistency loss and increased masking probability for MLM}
GRIT encourages similar examples to be grouped within each mini-batch, hence, the ITM\textsubscript{hard} can become more effective since the mini-batch may contain informative, hard negative samples. However, when GRIT is combined with ITC, one potential drawback is that the representations for similar samples would move away from each other unexpectedly, since all negatives will be equally penalized during the contrastive learning regardless of the similarity. 

To address this issue, we add a consistency loss that can reflect the similarity among samples. 
Namely, when an image $V$ and a text $T$ form a positive pair $(V,T)$, it is natural to assume that they share a similar semantic. Hence, we would expect the similarity scores $\bm{p}^{v2t}(V)$ and $\bm{p}^{t2v}(T)$ to be similar to each other. 
To this end, we define the soft pseudo-target $\tilde{\bm{p}}^{t2v}(T)$ as $sg(\bm{p}^{t2v}(T))$ and $\tilde{\bm{p}}^{v2t}(V)$ as $sg(\bm{p}^{v2t}(V))$ for $\bm{p}^{v2t}(V)$ and $\bm{p}^{t2v}(T)$, respectively, in which $sg(\cdot)$ is the stop-gradient operator. Then, our ITC with consistency loss is defined as
\begin{equation}
    \mathcal{L}_\text{ITC\textsubscript{cons}} = \mathcal{L}_\text{ITC} \, + \,  \frac{\lambda\textsubscript{cons}}{2}\mathbb{E}_{(V,T) \sim D} [ KL(\tilde{\bm{p}}^{v2t}(V)\,||\,\bm{p}^{t2v}(T)))+ KL(\tilde{\bm{p}}^{t2v}(T)\,||\,\bm{p}^{v2t}(V))],\label{eq:Consistency_loss}
\end{equation}
in which $\lambda\textsubscript{cons}$ is the regularization parameter. 
We expect this loss refines similarity scores which affect the quality of the grouping phase of GRIT. We set $\lambda\textsubscript{cons}$ as $0.2$ for all cases for simplicity. 

Finally, our model, dubbed as GRIT-VLP and illustrated in Fig \ref{fig:overall_process}, is obtained as follows. We use ALBEF\textsubscript{Base} as our base model architecture, and combine our GRIT, ITC consistency loss, and masking probability of 50\% for MLM. Consequently, the pre-training objective of GRIT-VLP is 
\begin{equation}
    \mathcal{L} = \mathcal{L}_\text{ITM\textsubscript{hard}}+\mathcal{L}_\text{MLM\textsubscript{50}}+\mathcal{L}_\text{ITC\textsubscript{cons}},
    \label{eq:total_loss}
\end{equation}
in which the mini-batches are generated by the GRIT strategy. The pseudo-code for GRIT-VLP is given in Alg.1/2 in the S.M.

\section{Experimental Results}\label{sec:experiments}
\subsection{Data and experimental settings}\label{subsec:data_exp}
\noindent{\textbf{[Training data]}} Following ALBEF \cite{(ALBEF)li2021align} and UNITER \cite{(UNITER)chen2020uniter}, we use four datasets (MS-COCO \cite{(MS-COCO)lin2014microsoft}, Visual Genome \cite{(VG)krishna2017visual}, Conceptual Captions \cite{(CC)sharma2018conceptual} and SBU Captions \cite{(SBU)ordonez2011im2text}) for training, which consist of 4M unique images and 5M image-text pairs.

\noindent{\textbf{[Implementation details]}}
Here, we give the concrete model architecture of ALBEF\textsubscript{Base}.
We use a 12-layer vision transformer ViT-B/16 \cite{(VIT)dosovitskiy2020image} with 86M parameters as the image encoder $f_v$ and initialize it with the weights pre-trained on ImageNet-1k \cite{(imagenet1K)touvron2021training}. 
A 6-layer Transformer \cite{(attention)vaswani2017attention} is used for both the text encoder $f_t$ and the multi-modal encoder $h$, which are initialized with the first 6 layers and the last 6 layers of BERT-base with 123.7M parameters\cite{(BERT)devlin2018bert}, respectively. 
We use the same data augmentation technique of ALBEF, and our model is trained for 20 epochs. All experiments are performed on $4$ NVIDIA A100 GPUs.
Furthermore, unless otherwise noted, we set $N=96$, $M=960$, and $L=48,000$. 
For all other hyper-parameter settings,  we follow ALBEF \cite{(ALBEF)li2021align}.
More details on the dataset, software platform, training procedures, and hyper-parameters are in the S.M.

\subsection{Downstream vision and language tasks} \label{sec: downstream}
After the pre-training step, our model is fine-tuned on three well-established downstream vision and language tasks, including image-text retrieval (IRTR), visual question answering (VQA2 \cite{(VQAv2)goyal2017making}), and natural language for visual reasoning (NLVR2 \cite{(NLVR)suhr2018corpus}). 
For IRTR, we use MS-COCO \cite{(MS-COCO)lin2014microsoft} and Flickr30K (F30K) \cite{(Flickr)plummer2015flickr30k} re-splited by \cite{(Flickr_split)karpathy2015deep}. 
We do not include SNLI-VE \cite{(SNLI-VE)xie2019visual} in the evaluation, since the data set is known to be noisy according to \cite{(e-snli)do2020snli}.
We mostly follow the fine-tuning and evaluation process of ALBEF \cite{(ALBEF)li2021align} except for using the momentum distillation.
We compare our method with various VLP methods trained on the same 4M training set. More details on the downstream tasks including evaluation setting are given in S.M. 

\begin{table}[!t]
\centering
\caption{Comparison with various methods on downstream vision-language tasks. \textbf{Bold} denotes the best result among models trained with 4M dataset.}
\label{tab:main_result}
\resizebox{\linewidth}{!}{%
\begin{tabular}{lcl|cccccccccccccccccccccccc} 
\hline
\multirow{2}{*}{Method} & \multirow{2}{*}{\begin{tabular}[c]{@{}c@{}}\#Pre-train\\~ Images\end{tabular}} & \multirow{2}{*}{}     &                      & \multicolumn{5}{c}{Flickr~R@1} &  & \multicolumn{5}{c}{COCO~R@1} &  &  & \multicolumn{3}{c}{VQA} &  &  &  & \multicolumn{3}{c}{NLVR2} &   \\
                        &                                                                                &                       & \multicolumn{1}{l}{} &  & TR   &  & IR   &               &  &  & TR            &  & IR   &   &  &  & test-dev &  & test-std  &  &  &  & dev   &  & test-P         &   \\ 
\hline
UNITER \cite{(UNITER)chen2020uniter}                  & 4M                                                                             &                       &                      &  & 87.3 &  & 75.6 &               &  &  & 65.7          &  & 52.9 &   &  &  & 72.70    &  & 72.91     &  &  &  & 77.18 &  & 77.85          &   \\
VILLA \cite{(VILLA)gan2020large}                   & 4M                                                                             &                       &                      &  & 87.9 &  & 76.3 &               &  &  & -             &  & -    &   &  &  & 73.59    &  & 73.67     &  &  &  & 78.39 &  & 79.30          &   \\
OSCAR \cite{(OSCAR)li2020oscar}                   & 4M                                                                             &                       &                      &  & -    &  & -    &               &  &  & 70.0          &  & 54.0 &   &  &  & 73.16    &  & 73.44     &  &  &  & 78.07 &  & 78.36          &   \\
ViLT \cite{(ViLT)kim2021vilt}                    & 4M                                                                             & \multicolumn{1}{c|}{} &                      &  & 83.5 &  & 64.4 &               &  &  & 61.5          &  & 42.7 &   &  &  & 71.26    &  & -         &  &  &  & 75.70 &  & 76.13          &   \\
ALBEF \cite{(ALBEF)li2021align}                   & 4M                                                                             &                       &                      &  & 94.3 &  & 82.8 &               &  &  & 73.1          &  & 56.8 &   &  &  & 74.54    &  & 74.70     &  &  &  & 80.24 &  & 80.50          &   \\

\hline
\textbf{GRIT-VLP\textsubscript{E-10}}             & 4M                                                                             &                       &                      &  & 94.7 &  & 82.0 &               &  &  & 74.9          &  & 58.1 &   &  &  & 74.72    &  & 74.74     &  &  &  & 79.98 &  & 80.11          &   \\
\textbf{GRIT-VLP}                 & 4M                                                                             &                       &                      &  & \textbf{96.0} &  & \textbf{83.8} &               &  &  & \textbf{77.1} &  & \textbf{59.5} &   &  &  & \textbf{75.11}    &  & \textbf{75.26}     &  &  &  & \textbf{80.73} &  & \textbf{81.60}          &   \\
\hline
ALBEF                   & 14M                                                                            &                       &                      &  & 95.9 &  & 85.6 &               &  &  & 77.6          &  & 60.7 &   &  &  & 75.84    &  & 76.04     &  &  &  & 82.55 &  & 83.14          &   \\ 
\hline

\end{tabular}
 }
\begin{spacing}{-0.2}
\end{spacing}
\end{table}
\vspace{-.1in}
\subsection{Comparison with the state-of-the-art VLP methods} \label{sec: overall_result}
Since we mainly build our method upon ALBEF, the previous state-of-the-art, we mainly compare our method with it.
Table \ref{tab:main_result} reports the results of GRIT-VLP with $N=128$ and  $M=1920$ on IRTR, VQA, and NLVR2.
In S.M, we present additional results on these hyper-parameters showing the robustness of our method with respect to $N$. \\
On  all downstream tasks (IRTR, VQA, NLVR2), GRIT-VLP outperforms other methods trained on the same 4M dataset, including the previous best model ALBEF (4M) by a large margin (+4\% TR/R@1 on MS-COCO, $+1.1\%$ on  NLVR test-P ). Moreover, GRIT-VLP is even competitive with ALBEF (14M) on some metrics, while being trained on a much smaller dataset. Furthermore,
``GRIT-VLP\textsubscript{E-10}'', denoting GRIT-VLP trained for only 10 epochs, achieves competitive performance compared to ALBEF (4M) trained with 30 epochs, highlighting the efficiency of our method. 
We believe the performance gains in Table \ref{tab:main_result} clearly highlights effectiveness of GRIT-VLP.

\begin{table}[!t]
\centering
\caption{Ablation study on the proposed method.}
\label{tab:ablation}
\resizebox{\linewidth}{!}{%

\begin{tabular}{c|lcc|c|c|l|cccccccccccccccc|c|c}
\hline
\multicolumn{4}{c|}{GRIT}  & \multicolumn{1}{l|}{\multirow{2}{*}{$\lambda_{cons}$}} & \multirow{2}{*}{\begin{tabular}[c]{@{}c@{}}Masking~\\Prob(\%)\end{tabular}} & \multirow{2}{*}{} & \multicolumn{1}{l}{} & \multicolumn{1}{l}{} & \multicolumn{1}{l}{TR} & \multicolumn{1}{l}{} & \multicolumn{3}{c}{(COCO)}                           & \multicolumn{1}{l}{} & IR            & \multicolumn{1}{l}{} & \multicolumn{1}{l}{} & \multicolumn{1}{l}{} &                      & \multicolumn{1}{l}{NLVR} &                      & VQA            &  & Time                           \\ 
\cline{1-4}
Collecting &  & Shuffle &  & \multicolumn{1}{l|}{}                        &                                                                             &                   & R@1                  & \multicolumn{1}{l}{} & R@5                    & \multicolumn{1}{l}{} & R@10          & \multicolumn{1}{l}{} & R@1           & \multicolumn{1}{l}{} & R@5           & \multicolumn{1}{l}{} & R@10                 & \multicolumn{1}{l}{} & \multicolumn{1}{l}{} & (test)                   & \multicolumn{1}{l}{} & (test-std)     &  & per epoch                      \\ 
\hline
\xmark          &  & \xmark       &  & 0                                        & 15                                                                          &                   & 73.8                 &                      & 92.3                   &                      & 96.5          &                      & 57.7          &                      & 82.5          &                      & 89.6                 &                      &                      & 80.37                    &                      & 74.70          &  & ~2h 27m                        \\
\xmark          &  & \xmark       &  & 0                                           & 50                                                                          &                   & 75.6                 &                      & 93.2                   &                      & 96.7          &                      & 58.8          &                      & 83.2          &                      & \textbf{90.1}        &                      &                      & 80.54                    &                      & 75.07          &  & ~2h 27m                        \\ 
\hline

\cmark  &  & \cmark      &  & 0                                          & 50                                                                          &                   & 76.4                 &                      & 93.6                   &                      & 96.7          &                      & \textbf{59.6} &                      & 83.3          &                      & \textbf{90.1}        &                      &                      & 81.32                    &                      & 75.14          &  & ~2h 30m                        \\ 

\cmark (naive) &  &\xmark     &  & 0                                           & 50                                                                          &                   & 76.8                 &                      & 93.6                   &                      & 96.8 &                      & \textbf{59.6} &                      & \textbf{83.4} &                      & 90.0        &                      &                      & 80.63                    &                      & 75.16          &  &  \multicolumn{1}{c}{3h}  \\
\cmark  &  & \xmark    &  &  0                                             & 50                                                                          &                   & 74.7                 &                      & 93.2                   &                      & 96.6          &                      & 58.6          &                      & 82.8          &                      & 89.7                 &                      &                      & 80.50                    &                      & 75.06          &  & ~2h 30m                        \\
\hline
\cmark &  & \cmark     &  & 0.2                                            & 15                                                                          &                   & 76.2                 &                      & 93.4                   &                      & 96.8          &                      & 59.0          &                      & 83.1          &                      & \textbf{90.1}        &                      &                      & 81.21                    &                      & 74.98          &  & ~2h 30m                        \\
\cmark &  & \cmark       &  & 0.2                                        & 50                                                                          &                   & \textbf{77.1}        &                      & \textbf{93.8}          &                      & \textbf{97.0} &                      & 59.5 &                      & \textbf{83.4}          &                      & 90.0                 &                      &                      & \textbf{81.43}           &                      & \textbf{75.30} &  & ~2h 30m                       \\
\hline
\end{tabular}
}
\begin{spacing}{0.1}
\end{spacing}
\end{table}
\vspace{-.1in}
\subsection{Ablation studies on the proposed method} \label{exp: ablation_result}
Table \ref{tab:ablation} shows the effectiveness of each proposed component: GRIT, ITC consistency loss, and enlarged masking probability ($15\% \rightarrow 50\%$) for MLM. 
First two rows indicate ALBEF\textsubscript{Base} and ALBEF\textsubscript{Base\textsubscript{50}} analyzed in Section \ref{sec:background}, respectively. 
By integrating the ALBEF\textsubscript{Base\textsubscript{50}} with GRIT-variants (row $3,4$),  we can verify that the performance is significantly improved.
However, in the case of ``naive'' implementation version of GRIT described in Section \ref{sec:methods} (row $4$), the training time is significantly increased as expected. 
We believe that competitive results of row 3 and 4 clearly demonstrate the need of all components of GRIT.
Moreover, if the both shufflings are removed from GRIT while collecting the mini-batches using the previous epoch (row $5$), its performance is severely degraded due to the vanishing randomness. 
The last row denotes our final GRIT-VLP (ALBEF\textsubscript{Base{\textsubscript{50}}} + GRIT + consistency); 
by adding the consistency loss from row $3$, we verify that the overall performance is increased. 
Furthermore, the gains of the last two rows compared to the first two rows show the combined effect of ``GRIT + consistency'' at two different mask sampling probabilities.

\begin{table}[!t]
    \begin{minipage}{.5\linewidth}
\caption{Effect of GRIT on ITC}
\label{tab:GMS_analysis}
\resizebox{\linewidth}{!}{
\begin{tabular}{llll|lclclclclclc} 
\hline
\multirow{2}{*}{\begin{tabular}[c]{@{}l@{}}\\\end{tabular}} & \multirow{2}{*}{Training tasks} & \multirow{2}{*}{} & \multirow{2}{*}{} & \multirow{2}{*}{} &       &  & TR    &  & \multicolumn{3}{c}{(COCO)} &  & IR    &  & \multicolumn{1}{l}{}  \\
                                                            &                                 &                   &                   &                   & R@1   &  & R@5   &  & R@10  &  & R@1                  &  & R@5   &  & R@10                  \\ 
\hline
                                                            & ITC~                            &                   &                   &                   & 59.7 &  & 84.8  &  & 91.9 &  & 43.2                &  & 72.2 &  & 82.0                 \\
                                                            & Queue-based ITC                       &                   &                   &                   & 60.3 &  & 85.2  &  & 92.2 &  & 43.5                 &  & 72.3 &  & 82.3                 \\
                                                            & ITC + GRIT                       &                   &                   &                   & 63.0 &  & 87.1 &  & \textbf{93.2} &  & 45.2                &  & 72.9 &  & 82.1                 \\
                                                             & ITC\textsubscript{cons} + GRIT                       &                   &                   &                   & \textbf{64.3} &  & \textbf{87.4} &  & \textbf{93.2} &  & \textbf{46.3}                &  & \textbf{73.7} &  & \textbf{82.6}                 \\
\hline
\end{tabular}
}
\begin{spacing}{0.1}
\end{spacing}
    \end{minipage}%
    \begin{minipage}{.5\linewidth}

\caption{Results on top of TCL \cite{(TCL)yang2022vision} }
\label{tab:TCL}
\resizebox{\linewidth}{!}{%
\begin{tabular}{l|cccccccc} 
\hline
         &      & TR   & \multicolumn{2}{c}{(COCO)} & IR   &      & NLVR2  & VQA         \\
Method   & R@1  & R@5  & R@10 & R@1                 & R@5  & R@10 & (test-P) & (test-std)  \\ 
\hline
TCL      & 75.6 & 92.8 & 96.7 & 59.0                & 83.2 & 89.9 & 81.33  & 74.92       \\
TCL+ours & \textbf{77.3} & \textbf{94.1} & \textbf{97.2} & \textbf{60.2}                & \textbf{83.7} & \textbf{90.0} & \textbf{81.52}  & \textbf{75.36}       \\
\hline
\end{tabular}
}

    \end{minipage}
\begin{spacing}{0.1}
\end{spacing}
\end{table}
\begin{figure*}[t]
    \centering
    \subfigure{
    \includegraphics[width=\linewidth]{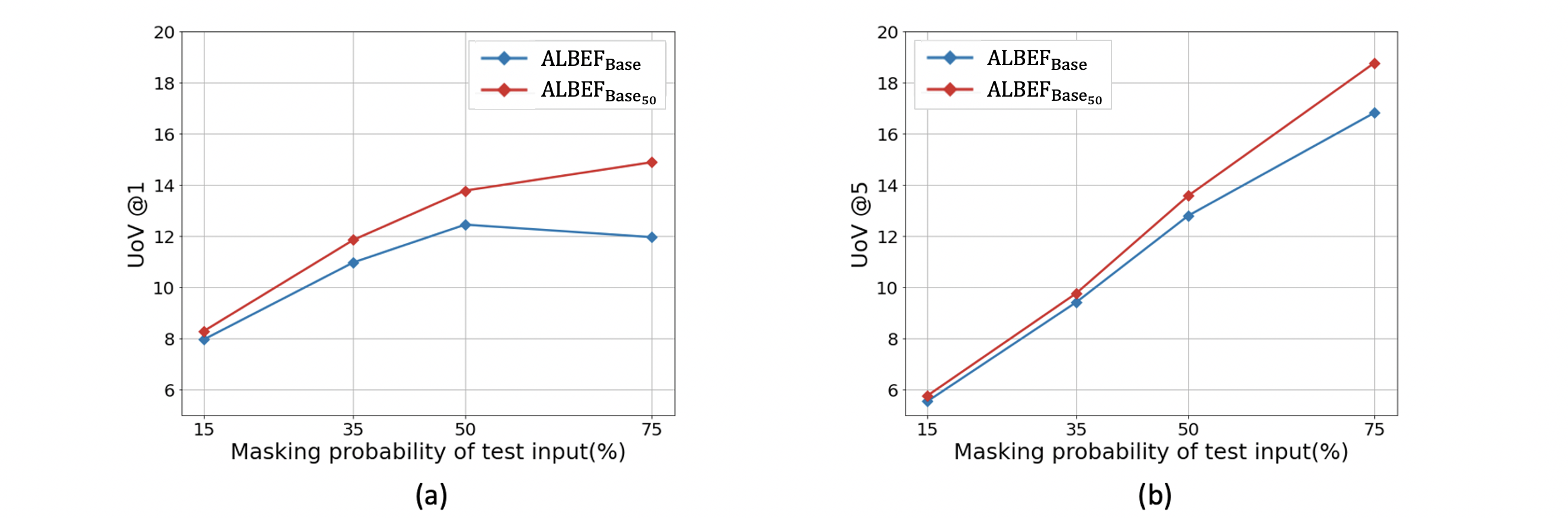}}
    \caption{UoV results on COCO validation dataset.
    }
    \label{fig:visual_literacy}
\begin{spacing}{0.1}
\end{spacing}
\end{figure*}
\vspace{-.1in}
\subsection{Experiments on the model-agnostic property}
To investigate the model-agnostic property of GRIT, we report the results when our method is integrated with different network architectures and objectives. \\
\noindent\textbf{[Small model, ITC]} Table \ref{tab:GMS_analysis} compares the IRTR results with several variants that are pre-trained and fine-tuned with only ITC loss and dual uni-modal encoders of ALBEF (without multi-modal encoder).
``Queue-based ITC'' denotes a model pre-trained with ITC and queues for leveraging the stored features from the momentum encoder (row $2$), and the other two models denote models that adopt our approach. 
While the performance gain of the Queue-based ITC is marginal, our GRIT brings a significant gain when combined with ITC. 
Finally, ITC\textsubscript{cons}+GRIT achieves the best result, demonstrating the standalone effectiveness of GRIT and consistency loss on ITC.
We believe this result shows that our method has a potential to be easily integrated with existing contrastive learning based models. 

\noindent\textbf{[Large model, more objectives]} In Table \ref{tab:TCL}, we additionally measure the gains of the recently proposed TCL \cite{(TCL)yang2022vision} when it is combined with ``ours'' (GRIT and enlarged masking probability). 
TCL introduces additional objectives and adopts almost the same but larger network architecture (additional momentum model) than ours. 
We omit the consistency loss since TCL already uses MD which has a similar role to it.  
Although the learning objectives and model sizes are different, we verify that the combination of our method and TCL again brings significant gains in Table \ref{tab:TCL}, which clearly shows the model-agnostic property of our method.
Details about this combined approach can be found in S.M.

\subsection{Analysis on the masking probabilities} \label{exp: mlm_analysis}
We believe the enlarged masking probability for MLM enables more usage of the visual features for predicting the masked token, resulting in a more effective multi-modal representation learning. To demonstrate this point, we introduce a \textit{Usage of Vision} (UoV) metric like \cite{(mlm_efficient)bitton2021data}, which is defined as the difference between MLM accuracy of a pre-trained model with and without the image input (\textit{Accuracy - Accuracy w/o image}). 
When evaluating the MLM \textit{Accuracy} and \textit{Accuracy w/o image} of the pre-trained model, test input sentences are masked with the same specific masking probability, and exactly the same tokens are masked for a fair comparison.
Then, a high UoV value means that the pre-trained model is highly affected by visual information since it implies that the vision information is important for the model to correctly predict the masked token. 

In Fig. \ref{fig:visual_literacy}, two pre-trained networks (ALBEF\textsubscript{Base}, ALBEF\textsubscript{Base\textsubscript{50}} ) are evaluated with test input sentences masked with various masking probability.
We calculate the UoV@1 and UoV@5 by considering the top-1 and top-5 MLM accuracy, respectively.
Fig. \ref{fig:visual_literacy} shows that ALBEF\textsubscript{{Base\textsubscript{50}}} model always obtain higher UoV demonstrating the high usage of vision information. In particular, when the textual context almost disappears (75\%), the difference in UoV becomes outright. 
As a result, we verify that enlarging the masking probability enriches the usage of visual information, which facilitates the alignment of image and text.

\section{Concluding Remarks}
We proposed GRIT-VLP, which effectively samples mini-batches for mining hard negatives while maintaining the computational overhead. 
We showed our method achieves state-of-the-art performance on various downstream tasks with much less computational overhead and can be easily integrated with existing VLP. 

\section*{\textcolor{\tsmcolor}{Acknowledgment}}
This work was supported in part by New Faculty Startup Fund from Seoul National University, NRF Mid Career Research Program [NRF-2021R1A2C2007884], IITP grants funded by the Korean government [No.2021-0-01696, High-Potential Individuals Global Training Program], [No.2021-0-01343, Artificial Intelligence Graduate School Program (Seoul National University)], [No.2021-0-02068, Artificial Intelligence Innovation Hub (Artificial Intelligence Institute, Seoul National University)], and SNU-NAVER Hyperscale AI Center.

\clearpage

\bibliographystyle{splncs04}
\bibliography{egbib}
\clearpage

\begin{center}
\textbf{\large Supplementary Material for \\ GRIT-VLP: Grouped Mini-batch Sampling \\for Efficient Vision and Language Pre-training}
\end{center}

\setcounter{equation}{0}
\setcounter{figure}{0}
\setcounter{table}{0}
\setcounter{section}{0}
\setcounter{subsection}{0}%
\setcounter{footnote}{0}%

\section{Data and Implementation Details}
Here we describe the details of software platform and dataset.
All experiments are conducted with four NVIDIA A100 GPUs. We use Python 3.7 and Pytorch \cite{(pytorch)paszke2017automatic} with CUDA 11.1 to implement GRIT-VLP. Table \ref{tab:data_statistics} summarizes the statistics of pre-training dataset. 

\begin{table}
\centering
\caption{Statistics of dataset}
\label{tab:data_statistics}
\begin{tabular}{c|ccccccccc} 
\toprule
          &  & COCO &  & VG   &  & SBU  &  & CC    &     \\ 
\hline
\# images &  & 113K &  & 100K &  & 851K &  & 2.82M &     \\
\# texts  &  & 567K &  & 769K &  & 851K &  & 2.82M &     \\
\bottomrule
\end{tabular}
\begin{spacing}{0.1}
\end{spacing}
\end{table}
\section{Details of Pre-training}
\subsection{Computation comparison}
We note that GRIT-VLP does not use the momentum encoder and momentum distillation in the pre-training phase. 
Table \ref{tab:computation} shows the computational costs when $N=96$ and $M=960$. 
As can be verified in Table \ref{tab:computation}, the model parameters, time per training one epoch, and queue size of GRIT-VLP are much smaller than the ALBEF, which clearly shows the efficiency of GRIT-VLP.

\begin{table}{
\centering
        \caption{Computational costs}
\label{tab:computation}
\begin{tabular}{c|c|c|c} 
\hline
Model  & Time per epoch & Parameters   & Queue  \\ 
\hline
ALBEF  & 3h 10m         & 420M   & 65536 \\
\hline
GRIT-VLP & 2h 30m         & 210M   & 48000  \\
\hline
\end{tabular}

}\end{table}

\subsection{Overall Process of GRIT-VLP}
Here, we provide a pseudo-code of the overall process of GRIT-VLP, mainly focusing on the GRIT we proposed. 
For simplicity, we omit the explanation about functions that are not related to the GRIT when we describe Algorithms (\ref{alg:overall},\ref{alg:grouping}).
In Algorithm \ref{alg:overall}, note that in the example-level shuffling (\textit{Phase 2}), all three queues containing a queue for image features, a queue for text features, and a queue for indices are shuffled in the same order to keep image-text pairs and their corresponding indices unmixed.
Likewise, the split procedure (\textit{Phase 2}) applies equally to all three queues.
As described in Algorithm \ref{alg:grouping}, in grouping phase (\textit{Phase 3}), we exclude previously visited indices to avoid including duplicate examples. After grouping phases are done for all the small sub-queues split from original queue, we make the original queue empty. 
Thus, each example in the dataset goes through the grouping phase only once per epoch.

\begin{algorithm}
        \caption{Entire process: Pseudocode, Pytorch-like}  \label{alg:overall}
        \scriptsize
        \PyComment{L\_Q\_v, L\_Q\_t - queues size of ($L \times d$) for saving feature representations }\\
        \PyComment{L\_Q\_idx - queues size of $L $ for saving unique indices }\\
        \PyComment{f\_v, f\_t - image encoder and text encoder }\\
        \PyComment{g\_v, g\_t - linear embbedings for image features and text features}\\
        \PyComment{I - index queue size of $M$ }\\ 
        \PyComment{G - output index array size of $D$; ($D = \#$(total examples)) }\\ 
        
        \PyComment{example\_shuffle(L\_Q\_v, L\_Q\_t, L\_Q\_idx) - a function that shuffles across the examples }\\ 
        \PyComment{divide(L\_Q\_v, L\_Q\_t, L\_Q\_idx) - a function for dividing each input queues} \\
        \PyComment{grouping(s\_Q\_v, s\_Q\_t, s\_Q\_idx) - a function for grouping (described in Algorithm \ref{alg:grouping})}\\ 
        \PyComment{cal\_itc\_cons\_loss(v\_feats, t\_feats) - a function for calculating ITC and consistency loss }\\ 
        \PyComment{mini\_batch\_shuffle(G) - a function that shuffles across the mini-batches }\\ 
    
        \PyCode{for e in epoch:} \\
        \PyCode{$\,\,\,\,\,\,\,\,\,\,$ loader $\leftarrow$ initialize(G)}
        \PyComment{Initialize indices of loader with $ \texttt{idxSet}_{e} $ }\\     
        \PyCode{$\,\,\,\,\,\,\,\,\,\,$ for batch\_i, (idx,V,T) in enumerate(loader):} 
        \PyComment{idx denotes the unique index of example }
        \\
         \PyCode  { $\,\,\,\,\,\,\,\,\,\,$ $\,\,\,\,\,\,\,\,$ }
         \PyComment{ITC and contrastive loss} \\
         \PyCode  { $\,\,\,\,\,\,\,\,\,\,$ $\,\,\,\,\,\,\,\,$ }
         \PyComment{Forwarding for uni-modal encoders } \\ 
         \PyCode  { $\,\,\,\,\,\,\,\,\,\,$ $\,\,\,\,\,\,\,\,\,\,$ v\_embeds = f\_v(V)  } \\
         \PyCode  { $\,\,\,\,\,\,\,\,\,\,$ $\,\,\,\,\,\,\,\,\,\,$ t\_embeds = f\_t(T)  } \\
         
         \PyCode  { $\,\,\,\,\,\,\,\,\,\,$ $\,\,\,\,\,\,\,\,$ } 
         \PyComment{Calculate [CLS] tokens for both image and text} \\
         \PyCode  { $\,\,\,\,\,\,\,\,\,\,$ $\,\,\,\,\,\,\,\,\,\,$ v\_feats = l2\_normalize(g\_v(v\_embeds[0]))  } 
         \PyComment{dim(v\_feats)= $N \times d$} \\ 
         \PyCode  { $\,\,\,\,\,\,\,\,\,\,$ $\,\,\,\,\,\,\,\,\,\,$ t\_feats = l2\_normalize(g\_t(t\_embeds[0]))  } 
          \PyComment{dim(t\_feats)= $N \times d$} \\ 

         \PyCode  { $\,\,\,\,\,\,\,\,\,\,$ $\,\,\,\,\,\,\,\,\,\,$ loss\_itc\_cons = cal\_itc\_cons\_loss(v\_feats, t\_feats)} 
         \PyComment{using (6) in manuscript} 
         \\
     
          \PyCode  { $\,\,\,\,\,\,\,\,\,\,$ $\,\,\,\,\,\,\,\,\,\,$ with torch.no\_grad():}  \\
          \PyCode { $\,\,\,\,\,\,\,\,\,\,$ $\,\,\,\,\,\,\,\,$ $\,\,\,\,\,\,\,\,\,\,$}
         \PyComment{Phase 1: Collection phase} \\
         \PyCode  { $\,\,\,\,\,\,\,\,\,\,$ $\,\,\,\,\,\,\,\,\,\,$  $\,\,\,\,\,\,\,\,\,\,$ L\_Q\_v.enqueue(v\_feats)  }\\
         \PyCode  { $\,\,\,\,\,\,\,\,\,\,$ $\,\,\,\,\,\,\,\,\,\,$  $\,\,\,\,\,\,\,\,\,\,$ L\_Q\_t.enqueue(t\_feats)  }\\
         \PyCode  { $\,\,\,\,\,\,\,\,\,\,$ $\,\,\,\,\,\,\,\,\,\,$ $\,\,\,\,\,\,\,\,\,\,$ L\_Q\_idx.enqueue(idx)  }\\
         
         \PyCode  { $\,\,\,\,\,\,\,\,\,\,$ $\,\,\,\,\,\,\,\,\,\,$  $\,\,\,\,\,\,\,\,\,\,$ if is\_full(L\_Q\_idx): } \\
         \PyCode { $\,\,\,\,\,\,\,\,\,\,$ $\,\,\,\,\,\,\,\,\,\,$ $\,\,\,\,\,\,\,\,\,\,$ $\,\,\,\,\,\,\,$ }
         \PyComment{$L$ features (indices) stored in queues} \\
         \PyCode { $\,\,\,\,\,\,\,\,\,\,$ $\,\,\,\,\,\,\,\,\,\,$ $\,\,\,\,\,\,\,\,\,\,$ $\,\,\,\,\,\,\,\,\,\,$}
         \PyComment{Phase 2: Example-level shuffle} \\
         \PyCode  { $\,\,\,\,\,\,\,\,\,\,$ $\,\,\,\,\,\,\,\,\,\,$ $\,\,\,\,\,\,\,\,\,\,$ $\,\,\,\,\,\,\,\,\,\,$ example\_shuffle(L\_Q\_v, L\_Q\_t, L\_Q\_idx) } \\
         \PyCode { $\,\,\,\,\,\,\,\,\,\,$ $\,\,\,\,\,\,\,\,\,\,$ $\,\,\,\,\,\,\,\,\,\,$ $\,\,\,\,\,\,\,\,\,\,$}
         \PyComment{Divide queues} \\ 
         \PyCode  { $\,\,\,\,\,\,\,\,\,\,$ $\,\,\,\,\,\,\,\,\,\,$ $\,\,\,\,\,\,\,\,\,\,$ $\,\,\,\,\,\,\,\,\,\,$  div\_Q\_v, div\_Q\_t, div\_Q\_idx = divide(L\_Q\_v, L\_Q\_t, L\_Q\_idx) } \\
         
         \PyCode { $\,\,\,\,\,\,\,\,\,\,$ $\,\,\,\,\,\,\,\,\,\,$ $\,\,\,\,\,\,\,\,\,\,$ $\,\,\,\,\,\,\,\,\,\,$}
        \PyComment{Phase 3: Grouping phase (Repeat for (L//M) times)} \\
        \PyCode  { $\,\,\,\,\,\,\,\,\,\,$ $\,\,\,\,\,\,\,\,\,\,$ $\,\,\,\,\,\,\,\,\,\,$ $\,\,\,\,\,\,\,\,\,\,$  for s\_Q\_v, s\_Q\_t, s\_Q\_idx in zip(div\_Q\_v, div\_Q\_t, div\_Q\_idx):} \\
        \PyCode { $\,\,\,\,\,\,\,\,\,\,$ $\,\,\,\,\,\,\,\,\,\,$ $\,\,\,\,\,\,\,\,\,\,$ $\,\,\,\,\,\,\,\,\,\,$ $\,\,\,\,\,\,\,\,\,\,$}
        \PyComment{$M$ features (indices) stored in small queues} \\
         \PyCode  { $\,\,\,\,\,\,\,\,\,\,$ $\,\,\,\,\,\,\,\,\,\,$ $\,\,\,\,\,\,\,\,\,\,$ $\,\,\,\,\,\,\,\,\,\,$ $\,\,\,\,\,\,\,\,\,\,$ I = grouping(s\_Q\_v, s\_Q\_t, s\_Q\_idx) } \\
        \PyCode  { $\,\,\,\,\,\,\,\,\,\,$ $\,\,\,\,\,\,\,\,\,\,$ $\,\,\,\,\,\,\,\,\,\,$ $\,\,\,\,\,\,\,\,\,\,$     $\,\,\,\,\,\,\,\,\,\,$ 
        G.append(I) } \\
        
        \PyCode { $\,\,\,\,\,\,\,\,\,\,$ $\,\,\,\,\,\,\,\,\,\,$ $\,\,\,\,\,\,\,\,\,\,$ $\,\,\,\,\,\,\,\,\,\,$}
        \PyComment{Make all three queues empty} \\
        \PyCode  { $\,\,\,\,\,\,\,\,\,\,$ $\,\,\,\,\,\,\,\,\,\,$ $\,\,\,\,\,\,\,\,\,\,$ $\,\,\,\,\,\,\,\,\,\,$  clear(L\_Q\_v, L\_Q\_t, L\_Q\_idx) } \\

        \PyCode  { $\,\,\,\,\,\,\,\,\,\,$ $\,\,\,\,\,\,\,\,$ }
        \PyComment{Calculate ITM and MLM loss (itm\_loss, mlm\_loss)} \\
        \PyCode  { $\,\,\,\,\,\,\,\,\,\,$ $\,\,\,\,\,\,\,\,$ }
        \PyComment{We omit the detailed procedures of ITM and MLM for simplicity} \\
        \PyCode  { $\,\,\,\,\,\,\,\,\,\,$ $\,\,\,\,\,\,\,\,$ }
        \PyComment{Note both losses require entire sequences of representations (v\_embeds, t\_embeds)} \\
        
        \PyCode  { $\,\,\,\,\,\,\,\,\,\,$ $\,\,\,\,\,\,\,\,\,\,$  loss=loss\_itc\_cons+loss\_itm+loss\_mlm} \\
        \PyCode  { $\,\,\,\,\,\,\,\,\,\,$ $\,\,\,\,\,\,\,\,\,\,$  loss.backward()} \\
        
        \PyCode { $\,\,\,\,\,\,\,\,$ }
        \PyComment{Phase 4: Mini-batch level shuffle} \\
        \PyCode{$\,\,\,\,\,\,\,\,\,\,$ mini\_batch\_shuffle(G)}
\end{algorithm} 
\setlength{\textfloatsep}{0pt}

\begin{algorithm}[!h]
        \caption{\textit{Phase 3:} Grouping phase: Pseudocode, Pytorch-like}  \label{alg:grouping}

         \scriptsize
        \PyComment{s\_Q\_v, s\_Q\_t - queues size of ($M \times d$) for saving feature representations }\\
        \PyComment{s\_Q\_idx - queues size of $M$ for saving unique indices }\\
        \PyComment{I - output index queue size of $M$ }\\ 
        \PyComment{cur\_index, pre\_index - index of current step, index of previous step  }\\    
        
         \PyComment{exclude\_index(matrix,index) - a function that sets both row and column of matrix corresponding to the index as $0$}\\    

        \PyCode  { \textbf{def} grouping (s\_Q\_v, s\_Q\_t, s\_Q\_idx): } \\

        \PyCode{$\,\,\,\,\,\,\,\,\,\,$}
        \PyComment{Calculate contrastive similarity}\\    
        \PyCode  { $\,\,\,\,\,\,\,\,\,\,$ similarity = s\_Q\_v @ s\_Q\_t } 
        \PyComment{ dim(similarity)= $M \times M$ }\\    
        \PyCode  { $\,\,\,\,\,\,\,\,\,\,$ P\_v2t = softmax(similarity, dim=1) } 
        \PyComment{ dim(P\_v2t)= $M \times M$ }\\    
        \PyCode  { $\,\,\,\,\,\,\,\,\,\,$ P\_t2v = softmax(similarity.t(), dim=1) } 
        \PyComment{ dim(P\_t2v)= $M \times M$ }\\    
        
        \PyCode{$\,\,\,\,\,\,\,\,\,\,$}
        \PyComment{Randomly sample one example}\\    
        \PyCode  { $\,\,\,\,\,\,\,\,\,\,$  M = len(s\_Q\_idx) } \\
        \PyCode  { $\,\,\,\,\,\,\,\,\,\,$  pre\_index = randint(0,M-1) } \\ 
        \PyCode  {  $\,\,\,\,\,\,\,\,\,\,$  I.enqueue(s\_Q\_idx[pre\_index]) } \\
        \PyCode  {  $\,\,\,\,\,\,\,\,\,\,$  use\_v2t = True } 
        \PyComment{For simplicity, we started with (image $\rightarrow$ text) direction}\\ 
        
        \PyCode{$\,\,\,\,\,\,\,\,\,\,$}
        \PyComment{Iteratively find index}\\    
        \PyCode  { $\,\,\,\,\,\,\,\,\,\,$  for i in range(M-1): } \\
        \PyCode{$\,\,\,\,\,\,\,\,\,\,$ $\,\,\,\,\,\,\,\,\,\,$}
        \PyComment{Use P\_v2t, P\_t2v alternatively}\\    
         \PyCode  { $\,\,\,\,\,\,\,\,\,\,$ $\,\,\,\,\,\,\,\,\,\,$ if (use\_v2t):} \\
         \PyCode  { $\,\,\,\,\,\,\,\,\,\,$  $\,\,\,\,\,\,\,\,\,\,$ $\,\,\,\,\,\,\,\,\,\,$ cur\_index=argmax(P\_v2t[pre\_index])}
         \PyComment{dim(P\_v2t[pre\_index])= $M$ }\\    
         \PyCode  { $\,\,\,\,\,\,\,\,\,\,$ $\,\,\,\,\,\,\,\,\,\,$ else:} \\
         \PyCode  { $\,\,\,\,\,\,\,\,\,\,$  $\,\,\,\,\,\,\,\,\,\,$ $\,\,\,\,\,\,\,\,\,\,$ cur\_index=argmax(P\_t2v[pre\_index])}
         \PyComment{dim(P\_t2v[pre\_index])= $M$ }\\    
         
         \PyCode{$\,\,\,\,\,\,\,\,\,\,$ $\,\,\,\,\,\,\,\,\,\,$ }
         \PyComment{Exclude pre\_index for both P\_v2t and P\_t2v whether use\_v2t is True or not.}\\    
         \PyCode  { $\,\,\,\,\,\,\,\,\,\,$ $\,\,\,\,\,\,\,\,\,\,$  exclude\_index(P\_v2t,pre\_index)}\\
         \PyCode  { $\,\,\,\,\,\,\,\,\,\,$ $\,\,\,\,\,\,\,\,\,\,$  exclude\_index(P\_t2v,pre\_index)}\\
         
         \PyCode  { $\,\,\,\,\,\,\,\,\,\,$ $\,\,\,\,\,\,\,\,\,\,$ I.enqueue(s\_Q\_idx[cur\_index])}\\
         \PyCode  { $\,\,\,\,\,\,\,\,\,\,$ $\,\,\,\,\,\,\,\,\,\,$ pre\_index = cur\_index}\\
        \PyCode  { $\,\,\,\,\,\,\,\,\,\,$  $\,\,\,\,\,\,\,\,\,\,$ use\_v2t = not use\_v2t  }\\
        
         \PyCode  { $\,\,\,\,\,\,\,\,\,\,$  $\bm{\text{return}}$ I}
\end{algorithm} 
\setlength{\textfloatsep}{1pt}

\subsection{First training epoch with GRIT}
Since GRIT-VLP obtains the indices for the grouped mini-batches from the previous epoch, randomized indices are given to the model at the first training epoch, as expected.
But, if we assume another pre-trained network for calculating similarity is available, we can generate indices considering the grouped mini-batches.
Then, we can use indices for the grouped mini-batches at the first epoch of the training. 
In our experiments, all GRIT-variant models use the generated indices from ALBEF\textsubscript{Base\textsubscript{50}} (pre-trained for one single epoch) at the first training epoch. 
Note that the effect of using this simple indices generation procedure at the first epoch is marginal in our experimental setting. 
But, we believe that the impact of this simple trick can be large when the training epoch is set \textit{very} small (\eg, one or two epoch).

\section{Details of Downstream Tasks}
As described in Section 5 (manuscript), we mainly follow the implementation details of ALBEF to fine-tune the pre-trained model. 
Unlike pre-training, we use randomly cropped image of resolution $384 \times 384$ in fine-tuning, and resize the images without cropping in inference.
The same RandAugment, optimizer, cosine learning rate decay, and weight decay are applied to all downstream tasks.
However, unlike ALBEF, since we do not use a momentum encoder in the pre-training phase, the momentum distillation (MD) is not utilized in all downstream tasks except for the Table 6 (manuscript) to show the model-agnostic property of our method. The total mini-batch size in S.M refers to the overall mini-batch size. Namely, it denotes ``number of GPUs $\times$ mini-batch size per GPU'' ($4 \times N$).  \\

\noindent{\textbf{[Image-Text Retrieval (IRTR)]}} 
IRTR finds the most similar text to a given image in a set of texts, or vice versa. 
In IRTR, with an updated momentum encoder during the pre-training phase, ALBEF \cite{(ALBEF)li2021align} fine-tunes the pre-trained model using queue-based ITC, MD for ITC, and ITM\textsubscript{hard}. 
To fairly evaluate the effectiveness of our method on the pre-training only, we mostly follow the fine-tuning phase of ALBEF by adopting the queue-based ITC (queue size: $65280$) and ITM\textsubscript{hard} (but, not adopting the MD for ITC).
Since we do not have an updated momentum encoder during the pre-training, we use the initial model (initialized with original BERT-base and ViT-B/16 pre-trained on ImageNet-1k) for initializing the momentum encoder. 
As another option, the momentum encoder can also be initialized with weights from the pre-trained model.
We empirically verify that the performance of these two variants is almost similar. 
Note that we use the momentum encoder and the additional queue only for constructing negative sets of queue-based ITC objective, not for the momentum distillation.
Although our method (GRIT and consistency loss) can also be incorporated into fine-tuning step for IRTR, we do not include them for a fair comparison. Unless otherwise noted, this fine-tuning setting for IRTR is applied in all experiments.

The train/validation/test set consists of 113k/5k/5k and 29k/1k/1k for COCO and F30K, respectively. 
In fine-tuning, we set the total batch size as $256$ and the initial learning rate as $1e-5$ for both datasets, and fine-tune for $5$ epochs for COCO, and $10$ epochs for F30K.
In evaluation, we first get the top-\textit{k} candidates based on image-text contrastive similarity. Then, we re-rank these calculated top-\textit{k} candidates using ITM scores. \textit{k} is set as 256 for COCO and 128 for F30K.  
\\

\noindent{\textbf{[Visual Reasoning ($\text{NLVR}2$)]}} 
NLVR classifies whether a textual description is true based on two images. The multi-modal encoder is consecutively duplicated to infer two images like \cite{(ALBEF)li2021align}. 
Since the model architecture has changed, one more pre-training is performed. Then, the pre-trained model is fine-tuned and evaluated on the NLVR2 dataset. 

We use the original train/val/test split of NLVR2\cite{(NLVR)suhr2018corpus} for evaluating visual reasoning.
Since NLVR tries to classify whether a textual description is true based on two images, the multi-modal encoder is duplicated to consider two images. 
As we mentioned above, since the overall structure of the model is changed, one more pre-training is performed with text-assignment task like ALBEF \cite{(ALBEF)li2021align}. Text-assignment task assigns the text to one of three choices: the first image, the second image, or none of them (three-way classification task). 
For an additional $1$ epoch NLVR pre-training step, we use images of size 256 × 256 on the 4M dataset with a total batch size of $256$ and a learning rate of $2e-5$.
After that, we fine-tune for $10$ epochs with a total batch size of $64$ and an initial learning rate of $2e-5$. 
We measure the performance on dev and test-P splits. 
\\

\noindent{\textbf{[Visual Question Answering (VQA)]}} VQA aims to derive an answer given an image and a relevant question. Following \cite{(ALBEF)li2021align}, an auto-regressive transformer decoder is added to generate answers. The decoder is initialized with weights of the pre-trained multi-modal encoder and fine-tuned with conditional language modeling loss. 
For VQA, we conduct experiment on the VQA2.0 dataset \cite{(VQAv2)goyal2017making}, where train/val/test split set is composed of 83k/41k/81k. 
Both training and validation sets are used for training, and also include additional question-answer pairs from Visual Genome, following previous works \cite{(LXMERT)tan2019lxmert,(ALBEF)li2021align}.
We fine-tune for $8$ epochs with a total batch size of $128$ and an initial learning rate of $2e-5$.
During inference, the decoder is constrained to only generate from the $3192$ candidate answers \cite{(Bilinear)kim2018bilinear} for a fair comparison.
We measure performance on the test-dev (t-dev) and test-std (t-std) splits. 

\section{Detailed explanations on Section 5.5 (manuscript)}
\subsection{Small model, ITC}
Note that all the models in Table 5 (manustript) are trained with dual uni-modal encoders. Thus, other losses like MLM and ITM can not be used.  
For the Queue-based ITC, queue size for storing extra negatives is set as $65280$. All other models are trained with \textit{in-batch} ITC. We verified that adding ITC\textsubscript{cons} and GRIT has considerable gain. We believe these results demonstrate the effectiveness of our method. 

\subsection {Large model, more objectives}
 We also further measure the gains when another type of VLP model, TCL \cite{(TCL)yang2022vision}, is combined with our method. TCL extends ALBEF with two complementary objectives in the pre-alignment step. They utilize the momentum encoder to design a informative pseudo-target of loss functions in the pre-training. Thus, TCL has a larger network architecture than ours (which do not contain momentum encoders in the pre-training) and is pre-trained with additional objectives. Despite these differences, we verify that the combination of our method and TCL again brings significant gains in Table 6 (manuscript). 
 
 For a fair comparison, we mostly follow the pre-training and fine-tuning settings of TCL. 
 Namely, we maintain the original architecture and objectives of TCL, and then additionally apply our grouped mini-batch sampling strategy and enlarged masking probability. Note that we do not employ the consistency loss, since TCL already has objectives for elaborating contrastive learning. 
 Since TCL uses a momentum encoder and distillation in the pre-training, for fine-tuning, we use an updated momentum encoder in the pre-training phase and MD in this experiment.

\section{Detailed Experimental Results} \label{sec: Detailed results}
Here we report additional results about analysis on sampling for VLP in Section 3 (manuscript). 
We include VQA results in both Table \ref{tab:itm_analysis_sup} and Table \ref{tab:mlm_analysis_sup}.
Moreover, we report models trained with various masking probabilities ($15\%, 35\%, 50\%, 75\%$) in Table  \ref{tab:mlm_analysis_sup}. 
\begin{table}
\centering
\caption{Ablation study on pre-training objectives of ALBEF on two V+L downstream tasks. MLM: masked language modeling with 15\% masking probability. }
\label{tab:itm_analysis_sup}
\resizebox{\linewidth}{!}{%
\begin{tabular}{lclllll|lclclclclclcllclcllcccl} 
\toprule
\multirow{2}{*}{\begin{tabular}[c]{@{}l@{}}\\~\\~\\\end{tabular}} & \multicolumn{1}{l}{\multirow{2}{*}{Epochs}} & \multirow{2}{*}{} & \multirow{2}{*}{} & \multirow{2}{*}{Training tasks} & \multirow{2}{*}{} & \multirow{2}{*}{} & \multirow{2}{*}{} &                &  & TR             &  & \multicolumn{3}{c}{(COCO)}         &  & IR             &  & \multicolumn{1}{l}{} &  & \multicolumn{1}{c}{} & \multicolumn{3}{c}{NLVR}             &  & \multicolumn{1}{c}{} & \multicolumn{3}{c}{VQA}                                                          & \multicolumn{1}{c}{}  \\
                                                                  & \multicolumn{1}{l}{}                        &                   &                   &                                 &                   &                   &                   & R@1            &  & R@5            &  & R@10           &  & R@1            &  & R@5            &  & R@10                 &  &                      & (dev)           &  & (test-P)          &  &                      & \multicolumn{1}{l}{(t-dev)} & \multicolumn{1}{l}{} & \multicolumn{1}{l}{(t-std)} &                       \\ 
\hline
                                                                  & \multirow{4}{*}{10}                         & \multirow{4}{*}{} & \multirow{4}{*}{} & MLM + ITM\textsubscript{rand}                       &                   &                   &                   & 61.6           &  & 86.1           &  & 92.5           &  & 47.8           &  & 75.4           &  & 84.8                 &  &                      & 77.02           &  & 78.44           &  &                      & 72.82                       &                      & 73.11                       &                       \\
                                                                  &                                             &                   &                   & MLM + ITM\textsubscript{rand} + ITC                 &                   &                   &                   & 66.8           &  & 88.8           &  & 94.5           &  & 51.1           &  & 78.4           &  & 86.8                 &  &                      & 76.59           &  & 78.69           &  &                      & 73.24                       &                      & 73.44                       &                       \\
                                                                  &                                             &                   &                   & MLM + ITM\textsubscript{hard}               &                   &                   &                   & 68.6           &  & 89.4           &  & 94.9           &  & 52.1           &  & 79.0           &  & 87.1                 &  &                      & 79.18 &  & 79.32           &  &                      & 73.55                           &                      & 73.67                           &                       \\
                                                                  &                                             &                   &                   & ALBEF\textsubscript{Base}                            &                   &                   &                   & \textbf{72.3} &  & \textbf{91.3} &  & \textbf{96.0} &  & \textbf{55.1} &  & \textbf{81.0} &  & \textbf{88.5}       &  &                      & \textbf{79.21}           &  & \textbf{79.78} &  &                      & \textbf{73.97}              &                      & \textbf{74.10}              &                       \\ 
\hline
                                                                  & \multirow{4}{*}{20}                         & \multirow{4}{*}{} & \multirow{4}{*}{} & MLM + ITM\textsubscript{rand}                       &                   &                   &                   & 66.5           &  & 88.3           &  & 94.0           &  & 51.3           &  & 78.3           &  & 86.5                 &  &                      & 78.02           &  & 79.43           &  &                      & 73.59                       &                      & 73.80              &                       \\
                                                                  &                                             &                   &                   & MLM + ITM\textsubscript{rand} + ITC                 &                   &                   &                   & 69.6           &  & 90.9           &  & 95.3           &  & 53.8           &  & 80.0           &  & 87.8                 &  &                      & 77.61           &  & 79.43           &  &                      & 73.68                       &                      & 73.98                       &                       \\
                                                                  &                                             &                   &                   & MLM + ITM\textsubscript{hard}        &                   &                   &                   & 72.0           &  & 91.5           &  & \textbf{96.6}           &  & 57.5           &  & 81.2           &  & 88.4                 &  &                      & \textbf{80.44}  &  & \textbf{80.83}  &  &                      & 74.19                       &                      & 74.43                       &                       \\
                                                                  &                                             &                   &                   & ALBEF\textsubscript{Base}                            &                   &                   &                   & \textbf{73.8}  &  & \textbf{92.3}  &  & 96.5  &  & \textbf{57.7}  &  & \textbf{82.5}  &  & \textbf{89.6}        &  &                      & 79.22           &  & 80.37           &  &                      & \textbf{74.62}              &                      & \textbf{74.70}                       &                       \\
\bottomrule
\end{tabular}
}
\begin{spacing}{0.1}
\end{spacing}
\end{table}

\subsection{Analysis on hard negative sampling}
Table \ref{tab:itm_analysis_sup} shows three downstream task performances (including VQA) of models with and without each objective of ALBEF\textsubscript{Base}.
As mentioned in Section 3 (manuscript), the bottom two rows of each epoch in the Table \ref{tab:itm_analysis_sup} reaffirm that the ITM\textsubscript{hard} is the most essential component for achieving great performance quickly in all downstream tasks including VQA.

\subsection{Analysis on masking probabilities}

\begin{table}
\centering
\caption{Evaluation of the various masking probability methods on two downstream V+L tasks. Base$_n$: model with $n\%$ masking probability.}
\label{tab:mlm_analysis_sup}
\resizebox{\linewidth}{!}{%
\begin{tabular}{lclllll|llclclclclclcll|lclcl|ccccc} 
\toprule
\multirow{2}{*}{\begin{tabular}[c]{@{}l@{}}\\~\\~\\~\\~\\~\\\end{tabular}} & \multicolumn{1}{l}{\multirow{2}{*}{Epochs}} & \multirow{2}{*}{} & \multirow{2}{*}{} & \multirow{2}{*}{Training tasks} & \multirow{2}{*}{} & \multirow{2}{*}{} & \multirow{2}{*}{} & \multirow{2}{*}{} &                &                      & TR             &                      & \multicolumn{3}{c}{(COCO)}                             &  & IR             &                      & \multicolumn{1}{l}{} &                      & \multicolumn{1}{c}{}  &                      & \multicolumn{3}{c}{NLVR}                                            & \multicolumn{1}{l}{} &  & \multicolumn{3}{c}{VQA}                                      &                       \\
                                                                           & \multicolumn{1}{l}{}                        &                   &                   &                                 &                   &                   &                   &                   & R@1            &                      & R@5            & \multicolumn{3}{c}{R@10}                                     & R@1            &  & R@5            & \multicolumn{3}{c}{R@10}                                           & \multicolumn{1}{c}{}  &                      & (dev)           &                      & \multicolumn{1}{l}{(test-P)} & \multicolumn{1}{l}{} &  & \multicolumn{1}{l}{(t-dev)} &  & \multicolumn{1}{l}{(t-std)} & \multicolumn{1}{l}{}  \\ 
\hline
                                                                           & \multirow{4}{*}{10}                         & \multirow{4}{*}{} & \multirow{4}{*}{} & ALBEF\textsubscript{Base}                            &                   &                   &                   &                   & 72.3           &                      & 91.3           &                      & 96.0           &                      & 55.1           &  & 81.0           &                      & 88.5                 &                      &                       &                      & 79.21           &                      & 79.78                      &                      &  & 73.97                       &  & 74.10                       &                       \\
                                                                           &                                             &                   &                   & ALBEF\textsubscript{Base\textsubscript{35}}                          &                   &                   &                   &                   & 73.1           &                      & 91.6           &                      & 96.2           &                      & 56.7           &  & 81.9           &                      & 89.2                 &                      &                       &                      & \textbf{79.42}           &                      & 79.78                      &                      &  & \textbf{74.42}                       &  & 74.42                       &                       \\
                                                                           &                                             &                   &                   & ALBEF\textsubscript{Base\textsubscript{50}}                          &                   &                   &                   &                   & \textbf{73.4} & \multicolumn{1}{c}{} & \textbf{92.5} & \multicolumn{1}{c}{} & \textbf{96.4} & \multicolumn{1}{c}{} & \textbf{57.2} &  & \textbf{82.3} & \multicolumn{1}{c}{} & \textbf{89.4}       & \multicolumn{1}{c}{} & \multicolumn{1}{c|}{} & \multicolumn{1}{c}{} & \textbf{79.42} & \multicolumn{1}{c}{} & \textbf{79.87}            &                      &  & \textbf{74.42}                       &  & \textbf{74.61}                       &                       \\
                                                                           &                                             &                   &                   & ALBEF\textsubscript{Base\textsubscript{75} }                         &                   &                   &                   &                   & 72.6           & \multicolumn{1}{c}{} & 91.5           & \multicolumn{1}{c}{} & 96.1           & \multicolumn{1}{c}{} & 56.3           &  & 81.4           & \multicolumn{1}{c}{} & 89.0                 & \multicolumn{1}{c}{} &                       &                      & 79.24           &                      & 79.33                      &                      &  & 74.38                       &  & 74.42                       &                       \\ 
\hline
                                                                           & \multirow{4}{*}{20}                         & \multirow{4}{*}{} & \multirow{4}{*}{} & ALBEF\textsubscript{Base}                            &                   &                   &                   &                   & 73.8           &                      & 92.3           &                      & 96.5           &                      & 57.7           &  & 82.5           &                      & 89.6                 &                      &                       &                      & 79.22           &                      & 80.37                      &                      &  & 74.62                       &  & 74.70                       &                       \\
                                                                           &                                             &                   &                   & ALBEF\textsubscript{Base\textsubscript{35}}                          &                   &                   &                   &                   & 75.2           &                      & 93.1           &                      & 96.6           &                      & 58.7           &  & 82.9           &                      & 89.8                 &                      &                       &                      & \textbf{80.56}  &                      & 80.47                      &                      &  & 74.79                       &  & 75.00                       &                       \\
                                                                           &                                             &                   &                   & ALBEF\textsubscript{Base\textsubscript{50} }                         &                   &                   &                   &                   & \textbf{75.6}  &                      & \textbf{93.2}  &                      & \textbf{96.7}  &                      & \textbf{58.8}  &  & \textbf{83.2}  &                      & \textbf{90.1}        &                      &                       &                      & 80.41  &                      & \textbf{80.54}             &                      &  & \textbf{74.94}                       &  & \textbf{75.07}                       &                       \\
                                                                           &                                             &                   &                   & ALBEF\textsubscript{Base\textsubscript{75}}                         &                   &                   &                   &                   & 75.1           & \multicolumn{1}{c}{} & 92.9           & \multicolumn{1}{c}{} & 96.8           & \multicolumn{1}{c}{} & 58.1           &  & 82.7           & \multicolumn{1}{c}{} & 89.6                 & \multicolumn{1}{c}{} &                       &                      & 79.30           &                      & 79.69                      &                      &  & 74.86                       &  & 74.85                       &                       \\
\bottomrule
\end{tabular}
}
\begin{spacing}{0.1}
\end{spacing}
\end{table}
\begin{figure*}
    \centering
    \subfigure{
    \includegraphics[width=\linewidth]{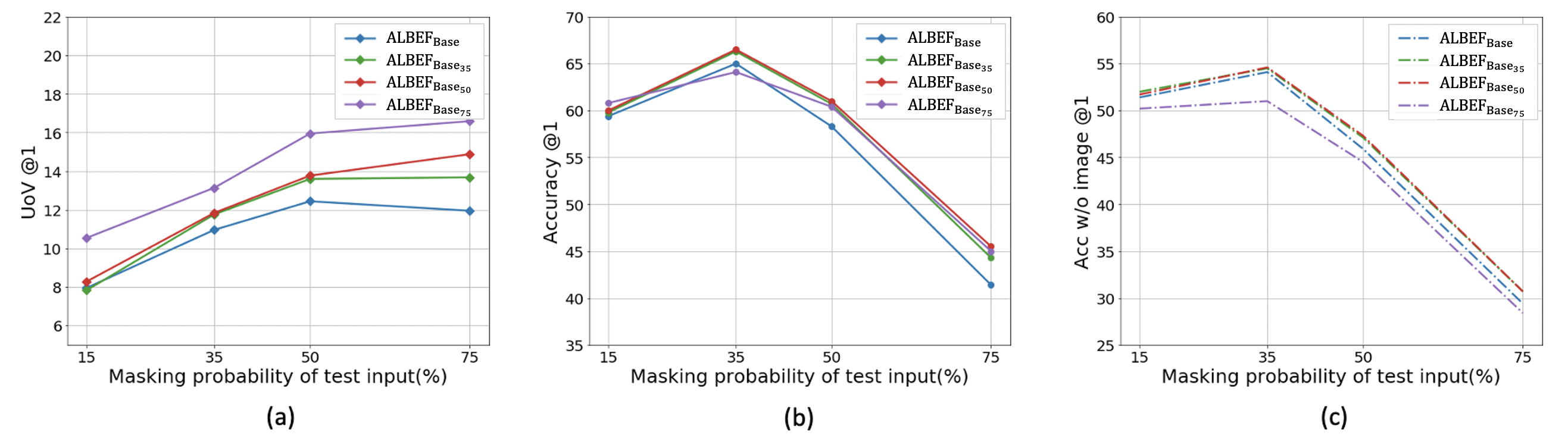}}
    \caption{Results of the four pre-trained models: (a) UoV (\textit{Usage of Vision}), (b) Accuracy and (c) Accuracy w/o image.}
    \label{fig:Visual literacy_sup}
\end{figure*}

Table \ref{tab:mlm_analysis_sup} shows the results of models trained with various masking probabilities (15\%, 35\%, 50\%, 75\%) on three downstream tasks including VQA. 
We verify that moderately enlarging the masking probability ($50\%$) enhances the final performance with a considerable gain.
However, in Table \ref{tab:mlm_analysis_sup}, we observe that if the masking probability is too high ($75\%$), the performance is rather degraded.
In this section, we scrutinize the reasons behind theses results with the MLM accuracy and UoV (\textit{Usage of Vision}). 

From the Fig. \ref{fig:Visual literacy_sup}(a), we observe that the UoV of the model trained with $75\%$ masking probability is significantly higher than that of other models. 
It means that the ALBEF\textsubscript{Base\textsubscript{75}} is the model that most reflects image information for prediction. 
However, as shown in the Table \ref{tab:mlm_analysis_sup}, the final performance of ALBEF\textsubscript{Base\textsubscript{75}} is lower than the ALBEF\textsubscript{Base\textsubscript{50}}. 
This result suggests that simply continuously increasing the usage of image information does not always lead to final performance improvement. 
To properly explain about this somewhat counter-intuitive result, it is necessary to consider the \textit{Accuracy w/o image} metric which indicates the MLM accuracy based only on the text information without image (Fig. \ref{fig:Visual literacy_sup}(c)), as well as the UoV metric (Fig. \ref{fig:Visual literacy_sup}(a)). 

Since the \textit{Accuracy w/o image} measures the MLM accuracy by using only text (without image) as input to the pre-trained model, it shows the ability of the pre-trained model to derive the correct answer for the masked word from the only textual information. 
In Fig. \ref{fig:Visual literacy_sup}(a) and Fig. \ref{fig:Visual literacy_sup}(c), we observe that the model pre-trained with sparse sentence ($75\%$) highly use the visual information, but its \textit{Accuracy w/o image} is relatively low compared to other models, which means that ALBEF\textsubscript{Base\textsubscript{75}} lacks the ability to leverage textual information.
We can observe more clearly that ALBEF\textsubscript{Base\textsubscript{75}} lacks the ability to utilize textual information in the NLVR2 task, which requires complex reasoning over sentences. 
As shown in the Table \ref{tab:mlm_analysis_sup}, the results of the ALBEF\textsubscript{Base\textsubscript{75}} on NLVR2 is even lower than the ALBEF\textsubscript{Base}. 
In conclusion, we verify that the moderately enlarged masking probability ($50\%$), which can utilize both visual and textual information in a balanced way, is more suitable for vision and language pre-training.

\begin{table}
\centering
\caption{Ablation studies on mini-batch size and search space}
\label{tab:batchsize}
\begin{tabular}{lcl|lcl|lccccccl|lcc} 
\cmidrule[\heavyrulewidth]{2-17}
 &                                                              &  &  &                                                                &  &  &      & TR   & \multicolumn{2}{c}{(COCO)} & IR   &      &  &  & \multicolumn{2}{c}{NLVR}  \\
 & \begin{tabular}[c]{@{}c@{}}$N$\end{tabular} &  &  & \begin{tabular}[c]{@{}c@{}}$M$ \end{tabular} &  &  & R@1  & R@5  & R@10 & R@1                 & R@5  & R@10 &  &  & (dev) & (test-P)          \\ 
\cmidrule{2-17}
 & 48                                                           &  &  & 480                                                            &  &  & 76.4 & 93.5 & 96.7 & 59.6                & 83.4 & \textbf{90.1} &  &  & 80.28 & 80.50             \\
 & 48                                                           &  &  & 960                                                            &  &  & 76.9 & \textbf{94.0} & \textbf{97.1} & 59.7                & \textbf{83.6} & \textbf{90.1} &  &  & 81.23 & 81.07             \\
 & 48                                                           &  &  & 1920                                                           &  &  & 76.8 & 93.7 & 96.9 & \textbf{60.0}                & 83.3 & 89.9 &  &  & 81.29 & 81.34             \\ 
\cmidrule{2-17}
 & 96                                                           &  &  & 480                                                            &  &  & 76.5 & 93.9 & 97.0 & 59.4                & 83.4 & 90.0 &  &  & 80.68 & 80.68             \\
 & 96                                                           &  &  & 960                                                            &  &  & \textbf{77.1} & 93.8 & 97.0 & 59.5                & 83.4 & 90.0 &  &  & \textbf{81.30} & 81.43             \\
 & 96                                                           &  &  & 1920                                                           &  &  & \textbf{77.1} & 93.5 & 96.9 & 59.6                & 83.5 & 90.0 &  &  & 80.53 & 81.01             \\ 
\cmidrule{2-17}
 & 128                                                          &  &  & 480                                                            &  &  & 76.3 & 93.7 & 96.9 & 59.1                & 83.1 & 89.7 &  &  & 80.73 & 80.88             \\
 & 128                                                          &  &  & 960                                                            &  &  & 77.0 & 93.5 & \textbf{97.1} & 59.6                & 83.3 & 90.0 &  &  & 80.54 & 81.47             \\
 & 128                                                          &  &  & 1920                                                           &  &  & \textbf{77.1} & 93.6 & 96.7 & 59.5                & 83.3 & 89.9 &  &  & 80.73 & \textbf{81.60}             \\
\cmidrule[\heavyrulewidth]{2-17}
\end{tabular}

\end{table}

\subsection{Ablation study on hyper-parameters}
Regarding the batch size ($N$) and search space size ($M$), we carry out an additional ablation study in Table \ref{tab:batchsize} (with other hyper-parameters fixed). We experiment with three mini-batch sizes ($48$, $96$, $128$) and search space sizes ($480, 960, 1920$). In all search space sizes, we verify that our method is generally robust with batch size. 
However, if the search space is small ($480$), the performance is degraded compared to the other search space size. If the search space is sufficiently large (more than $960$), the performance gap is marginal.
Namely, $N$ does not have a crucial impact on performance but $M$ does.
In Table 3 (manuscript), we select $N=128$ and $M=1920$ since the total batch size ($128 \times 4 = 512$) is same as ALBEF ($64 \times 8 = 512$) and it shows consistently great performance on all downstream tasks.

\begin{table}
\centering
\caption{Fine-tuned results image-text retrieval on Flickr30K and MSCOCO datasets}
\label{tab:IRTR}
\resizebox{\linewidth}{!}{%
\begin{tabular}{lcl|ccccccccccccc|ccccccccccccc} 
\toprule
\multirow{2}{*}{\begin{tabular}[c]{@{}l@{}}\\Method\end{tabular}} & \multirow{2}{*}{\begin{tabular}[c]{@{}c@{}}\#Pre-train\\~ Images\end{tabular}} & \multirow{2}{*}{} & \multicolumn{1}{l}{\multirow{2}{*}{}} & \multicolumn{12}{c|}{MSCOCO (5K test set)}                                                                                                                                                                                                                                                     & \multicolumn{13}{c}{Flickr30K (1K test set)}                                                                                                                                                                                                                                                                           \\
                                                                  &                                                                                &                   & \multicolumn{1}{l}{}                  & \multicolumn{1}{l}{}   & \multicolumn{1}{l}{} & TR                     & \multicolumn{1}{l}{} &                        & \multicolumn{1}{l}{} & \multicolumn{1}{l}{}   & \multicolumn{1}{l}{} & IR                     & \multicolumn{1}{l}{} & \multicolumn{1}{l}{}   & \multicolumn{1}{l|}{} & \multicolumn{1}{l}{} & \multicolumn{1}{l}{}   & \multicolumn{1}{l}{} & TR                     & \multicolumn{1}{l}{} & \multicolumn{1}{l}{}    & \multicolumn{1}{l}{} & \multicolumn{1}{l}{}   & \multicolumn{1}{l}{} & IR                     & \multicolumn{1}{l}{} & \multicolumn{1}{l}{}   & \multicolumn{1}{l}{}  \\ 
\midrule
                                                                  & \multicolumn{1}{l}{}                                                           &                   &                                       & R@1                    &                      & R@5                    &                      & R@10                   &                      & R@1                    &                      & R@5                    &                      & R@10                   &                       &                      & R@1                    &                      & R@5                    &                      & R@10                    &                      & R@1                    &                      & R@5                    &                      & R@10                   &                       \\
UNITER                                                            & 4M                                                                             &                   &                                       & 65.7                   &                      & 88.6                   &                      & 93.8                   &                      & 52.9                   &                      & 79.9                   &                      & 88.0                   &                       &                      & 87.3                   &                      & 98.0                   &                      & 99.2                    &                      & 75.6                   &                      & 94.1                   &                      & 96.8                   &                       \\
VILLA                                                             & 4M                                                                             &                   &                                       & -                      &                      & -                      &                      & -                      &                      & -                      &                      & -                      &                      & -                      &                       &                      & 87.9                   &                      & 97.5                   &                      & 98.8                    &                      & 76.3                   &                      & 94.2                   &                      & 96.8                   &                       \\
OSCAR                                                             & 4M                                                                             &                   &                                       & 70.0                   &                      & 91.1                   &                      & 95.5                   &                      & 54.0                   &                      & 80.8                   &                      & 88.5                   &                       &                      & -                      &                      & -                      &                      & -                       &                      & -                      &                      & -                      &                      & -                      &                       \\
ALBEF                                                             & 4M                                                                             &                   &                                       & 73.1                   &                      & 91.4                   &                      & 96.0                   &                      & 56.8                   &                      & 81.5                   &                      & 89.2                   &                       &                      & 94.3                   &                      & 99.4                   &                      & 99.8                    &                      & 82.8                   &                      & \textbf{96.7}                   &                      & \textbf{98.4}                   &                       \\ 
\midrule
\textbf{GRIT-VLP\textsubscript{E-10}}                                              & 4M                                                                             &                   &                                       & 74.9                   &                      & 93.0                   &                      & \textbf{97.0}                   &                      & 58.1                   &                      & 82.7                   &                      & 89.6                   &                       &                      & 94.7                   &                      & \textbf{99.6}                   &                      & \textbf{99.9}                    &                      & 82.0                   &                      & 95.3                   &                      & 97.7                   &                       \\
\textbf{GRIT-VLP}                                                  & 4M                                                                             &                   &                                       & \textbf{77.1}                   &                      & \textbf{93.6}                   &                      & 96.7                   &                      & \textbf{59.5}                   &                      & \textbf{83.3}                   &                      & \textbf{89.9}                   &                       &                      & \textbf{96.0}                   &                      & \textbf{99.6}                   &                      & \textbf{99.9}                    &                      & \textbf{83.8}                   &                      & 96.2                   &                      & 97.8                   &                       \\ 
\midrule
ALBEF                                                             & 14M                                                                            &                   &                                       & 77.6                   &                      & 94.3                   &                      & 97.2                   &                      & 60.7                   &                      & 84.3                   &                      & 90.5                   &                       &                      & 95.9                   &                      & 99.8                   &                      & 100.0                   &                      & 85.6 &                      & 97.5 &                      & 98.9 &                       \\
ALIGN                                                             & 1.8B                                                                           &                   &                                       & 77.0                   &                      & 93.5                   &                      & 96.9                   &                      & 59.9                   &                      & 83.3                   &                      & 89.8                   &                       &                      & 95.3                   &                      & 99.8                   &                      & 100.0                   &                      & 84.9                   &                      & 97.4                   &                      & 98.6                   &                       \\
\bottomrule
\end{tabular}
}
\end{table}

\subsection{Additional results on IRTR}
Table \ref{tab:IRTR} shows the full IRTR results including R@5 and R@10 accuracy on both COCO and Flickr30K dataset.
As we mentioned in the Section 5 (manuscript), GRIT-VLP (4M) outperforms other methods including ALBEF by a considerable gain, except for the IR results (R@5 and R@10) that are  slightly lower than ALBEF in the Flickr30K dataset.
In particular, in the COCO dataset, we verify that our GRIT-VLP (4M) shows competitive results with ALBEF (14M) and ALIGN (1.8B) trained on much larger datasets.
Moreover, when using the exact same dataset (4M), we observe that ``GRIT-VLP\textsubscript{E-10}'' (4M) trained with only $10$ epochs shows superior performance to the ALBEF (4M) trained with $30$ epochs, in the COCO dataset.
Note that our GRIT-VLP also obtains faster training time per epoch and smaller model parameters in the pre-training compared to ALBEF.
\clearpage

\end{document}